\PassOptionsToPackage{table}{xcolor}
\RequirePackage{fix-cm}
\documentclass{article}
\usepackage{iclr2027_conference,times}
\usepackage[letterpaper,left=1in,right=1in,top=0.8in,bottom=0.8in,
            headheight=0pt,headsep=0pt,footskip=24pt]{geometry}
\iclrfinalcopy
\makeatletter
\renewcommand{\normalsize}{\@setfontsize\normalsize{11pt}{14.5pt}\abovedisplayskip 11pt plus 2pt minus 2pt
  \belowdisplayskip 11pt plus 2pt minus 2pt
  \abovedisplayshortskip 5pt plus 2pt
  \belowdisplayshortskip 7pt plus 2pt minus 1pt}
\renewcommand{\small}{\@setfontsize\small{10pt}{13pt}}
\renewcommand{\footnotesize}{\@setfontsize\footnotesize{9.5pt}{12pt}}
\renewcommand{\scriptsize}{\@setfontsize\scriptsize{8pt}{10pt}}
\renewcommand{\tiny}{\@setfontsize\tiny{6.5pt}{8pt}}
\renewcommand{\large}{\@setfontsize\large{13pt}{17pt}}
\renewcommand{\Large}{\@setfontsize\Large{15pt}{19pt}}
\renewcommand{\LARGE}{\@setfontsize\LARGE{18pt}{22pt}}
\renewcommand{\huge}{\@setfontsize\huge{21pt}{25pt}}
\renewcommand{\Huge}{\@setfontsize\Huge{25pt}{30pt}}
\normalsize
\def\section{\@startsection{section}{1}{\z@}
  {-18pt plus -3pt minus -2pt}{9pt plus 2pt}
  {\fontsize{14}{18}\selectfont\scshape\raggedright}}
\def\subsection{\@startsection{subsection}{2}{\z@}
  {-14pt plus -2pt minus -1pt}{7pt plus 1pt}
  {\fontsize{12}{15.5}\selectfont\scshape\raggedright}}
\def\subsubsection{\@startsection{subsubsection}{3}{\z@}
  {-12pt plus -2pt minus -1pt}{6pt plus 1pt}
  {\normalsize\scshape\raggedright}}
\def\paragraph{\@startsection{paragraph}{4}{\z@}
  {9pt plus 2pt minus 1pt}{-0.65em}{\normalsize\bfseries}}
\renewenvironment{abstract}
  {\par\vspace{6pt}\begin{center}{\fontsize{12}{15}\selectfont\scshape Abstract}\end{center}
   \vspace{-3pt}\begingroup\normalsize\setlength{\parskip}{0pt}}
  {\par\endgroup\vspace{6pt}}
\long\def\author#1{\gdef\@author{#1}}
\def\@maketitle{\begingroup
  \centering
  {\fontsize{21}{26}\selectfont\bfseries\@title\par}
  \vspace{10pt}
  {\@author\par}
  \endgroup
  \vspace{8pt}}
\makeatother

\usepackage[T1]{fontenc}
\usepackage[mono,varqu]{zi4}
\usepackage{microtype}
\usepackage{amsmath,amssymb,booktabs,tabularx,array,longtable}
\usepackage{xcolor,caption,graphicx,float,enumitem,needspace,placeins}
\newcolumntype{R}{>{\raggedleft\arraybackslash}X}
\newcolumntype{L}{>{\raggedright\arraybackslash}X}
\newcolumntype{P}[1]{>{\raggedright\arraybackslash}p{#1}}
\newcommand{\PhysTableFont}{\fontsize{9.25}{12}\selectfont}
\newcommand{\PhysTableBefore}{\vspace{0pt}}
\newcommand{\PhysTableAfter}{\par\vspace{2pt}}
\newcommand{\PhysTableSetup}{\PhysTableFont
  \setlength{\tabcolsep}{1.4pt}\renewcommand{\arraystretch}{1.17}\setlength{\heavyrulewidth}{0.8pt}\setlength{\lightrulewidth}{0.4pt}\setlength{\aboverulesep}{2.4pt}\setlength{\belowrulesep}{2.4pt}\setlength{\LTcapwidth}{\linewidth}\arrayrulecolor{black}}
\newcommand{\PhysNote}[1]{\par\vspace{5pt}\noindent\begin{minipage}{\linewidth}\PhysTableFont #1\end{minipage}}
\newcommand{\AllTaskScore}{65.00}
\newcommand{\AllTaskSR}{58.57}
\newcommand{\DimensionScore}{68.14}
\newcommand{\DimensionSR}{62.00}
\newcommand{\FocalTaskScore}{75.06}
\newcommand{\FocalTaskSR}{65.00}
\newcommand{\TenTaskSR}{68.00}
\newcommand{\EightTaskScore}{62.88}
\newcommand{\EightTaskSR}{55.00}

\usepackage[hidelinks]{hyperref}
\usepackage{url}
\usepackage[nameinlink,noabbrev]{cleveref}
\hypersetup{pdftitle={PhysEvo: Astra Can Act, Let It},pdfauthor={Wenqing Tian, Zeyu Zhang, Zhaocheng Liu, Fengwei Liu, Qiang Liu, Liang Wang},pdfsubject={Physical recursive self-improvement for embodied control}}
\newcommand{\taskH}{H^{\mathrm{task}}}
\newcommand{\metaH}{H^{\mathrm{meta}}}
\newcommand{\tool}[1]{\texttt{#1}}
\title{PhysEvo: Astra Can Act, Let It}
\author{{\fontsize{12}{16}\selectfont
  Wenqing Tian\textsuperscript{1,2}\quad
  Zeyu Zhang\textsuperscript{1,2}\quad
  Zhaocheng Liu\textsuperscript{4,$\dagger$,$\ddagger$}\\[3pt]
  Fengwei Liu\textsuperscript{3}\quad
  Qiang Liu\textsuperscript{1,2,$\dagger$}\quad
  Liang Wang\textsuperscript{1,2}\par}
  \vspace{8pt}
  {\fontsize{10.5}{13.5}\selectfont
  \textsuperscript{1}University of Chinese Academy of Sciences\\
  \textsuperscript{2}Institute of Automation, Chinese Academy of Sciences\\
  \textsuperscript{3}Tsinghua University\qquad
  \textsuperscript{4}Independent Researcher\par}
  \vspace{5pt}
  {\fontsize{9.5}{12}\selectfont
  \href{mailto:lio.h.zen@gmail.com}{\texttt{lio.h.zen@gmail.com}}\qquad
  \href{mailto:qiang.liu@nlpr.ia.ac.cn}{\texttt{qiang.liu@nlpr.ia.ac.cn}}\par}
}

\makeatletter
\gdef\@thanks{\footnotetext[2]{Corresponding authors.}\footnotetext[3]{Project leader.}}
\makeatother

\begin{document}
\flushbottom
\maketitle
\begin{abstract}
Astra can act, yet reliable manipulation depends on the system through which it observes and controls the world. We introduce PhysEvo, a framework for physical recursive self-improvement (RSI) around a single frozen model. A task agent executes robot tasks; a meta-agent uses the resulting trajectories to diagnose failures, revise tools and skills, and test corrections. The meta-agent can also improve its own diagnostic tools, so retained revisions support both later action and later self-improvement. This process develops joint-level control, evidence-seeking observation, and reusable manipulation skills without model-weight updates or a separately trained action policy. Across 42 RoboDojo tasks, held-out-layout evaluation of retained task-specific deployment versions yields a five-dimension average score of \DimensionScore{}/100 and \DimensionSR{}\% success, compared with 47.17\% for RoboDawn's one-shot Astra agent, the strongest published reference in our comparison. On eight manipulation tasks challenging direct Astra, PhysEvo achieves \EightTaskSR{}\% success, compared with 1.25\% for the direct-Astra reference. Deploying the simulation-evolved harness on AgileX PiPER and continuing skill revision yields 90.60/100 average score and 84.00\% success across 25 trials on five real-world tasks. PhysEvo turns the consequences of action into persistent, testable changes to how a frozen model acts and improves.
\end{abstract}
\begin{figure}[H]
\centering
\includegraphics[width=\linewidth]{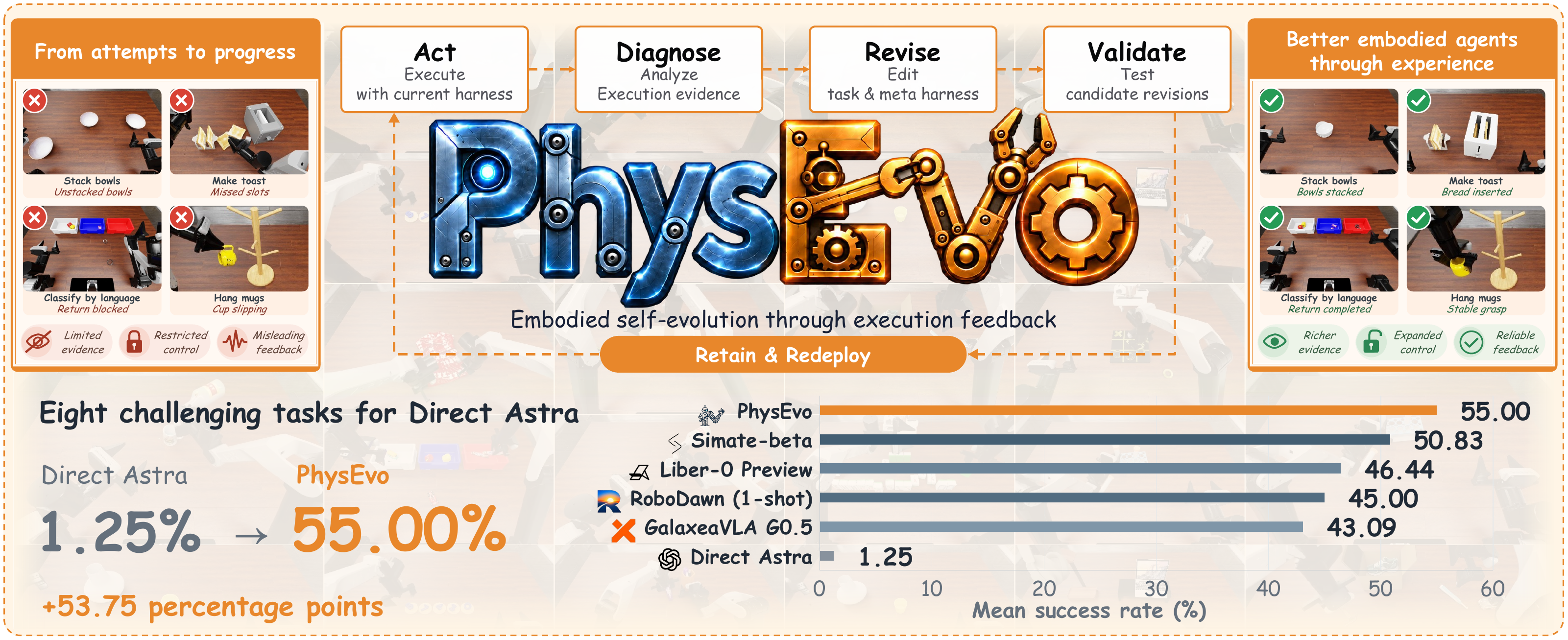}
\caption{Physical RSI in PhysEvo: execute, diagnose, revise, test, and inherit. The lower panel compares success on eight challenging tasks for Direct Astra (\cref{tab:f2eight}).}
\label{fig:overview}
\end{figure}
\clearpage
\section{Introduction}
\label{sec:introduction}
Astra can already control a robot, but demonstrated competence does not guarantee reliable execution. In a ten-task RoboDojo evaluation, direct Astra reaches 100\% success on object classification yet succeeds on only 26\% of attempts overall. Pairing it with the specialized vision--language--action (VLA) policy $\pi_{0.5}$ raises success to 48\% \citep{su2026astra}. These results pose a question: \emph{what prevents capabilities that are already present from being used reliably?}

Our executions expose bottlenecks in the \emph{harness}: the tools, skills, and execution logic between the model and the robot. During pouring, the held bottle obscures the wrist view needed to monitor liquid delivery. With a tool-center-point (TCP) interface, local inverse kinematics (IK) can fold the arm into an obstructive posture, while leaving Astra no explicit way to choose the joint configuration (\cref{fig:case_control,fig:case_observation}). These failures call for more than a new action through an unchanged interface: the agent needs a useful view or direct authority over posture.

Why should the same model be able to fix a failure it did not prevent? \emph{Retrospective diagnosis has access to evidence that was unavailable during action selection.} A trajectory reveals the consequences of a command; aligned images, robot states, and code help trace those consequences to a cause. Astra can then propose a concrete intervention, such as maintaining a free-arm side view or exposing joint reconfiguration, and test it through further execution. In our pouring case, revised skill guidance changes the observation strategy, and the subsequent episode succeeds. Physical interaction thus provides both evidence for a correction and a test of whether it works.

A correction need not end with the episode. Retained tools and skills change later execution; retained diagnostic tools change how later failures can be investigated. In our packing case, Astra reuses a review tool it built earlier to trace box displacement to preceding contact and revise clearance guidance (\cref{fig:case_diagnosis}). We call this loop \emph{physical recursive self-improvement} (RSI): experience revises the system used for action and diagnosis, and the revised system is inherited by subsequent rounds.

We introduce \textbf{PhysEvo}, which implements this loop with one frozen Astra in two roles. The task agent observes and controls the robot. The meta-agent inspects execution, edits tools and skills, tests revisions, and retains useful changes, including changes to its own diagnostic resources. The resulting framework combines joint-level control, active and historical observation, geometry tools, and manipulation skills.

Across 42 RoboDojo tasks, held-out-layout evaluation of retained task-specific deployment versions achieves \textbf{\DimensionScore{}/100 score and \DimensionSR{}\% success} under equal five-dimension weighting. This exceeds the 47.17\% reported by RoboDawn's one-shot Astra agent, the strongest reference in our comparison (\cref{tab:dimensions}) \citep{robodawn}. On the ten-task comparison with direct and hybrid policies, PhysEvo reaches \TenTaskSR{}\% success versus 26.00\% and 48.00\%, respectively. On eight manipulation tasks challenging direct Astra, it reaches \textbf{\EightTaskSR{}\% versus 1.25\%}, a \textbf{53.75-point gain}; Simate-beta, the strongest robot-policy reference on this subset, reports 50.83\% (\cref{fig:overview,tab:f2eight}) \citep{robodojo_board_20260923,zhang2026gpt6astrarobodojo}.

On AgileX PiPER, the simulation-evolved harness serves as a starting point for continued adaptation. Across five trials on each of five real-world tasks, PhysEvo achieves a mean score of 90.60/100 and 84.00\% success. Revisions compensate for liquid-flow effects and replace geometry-first localization with a visually guided approach followed by close-range geometric refinement (\cref{sec:hardware}). The same loop can therefore revise both how the robot moves and when it seeks geometric evidence.

\Needspace{6\baselineskip}
Our contributions are \textbf{(i)} a physical RSI framework in which one frozen model acts, revises, and inherits task and meta resources; \textbf{(ii)} an evolved interaction architecture combining joint control, evidence-seeking observation, and reusable skills; and \textbf{(iii)} a 42-task evaluation with mechanism-focused cases and a five-task evaluation of continued adaptation on real hardware.

\section{Related Work}
\label{sec:related}
\paragraph{Foundation models for robot control.}
Vision--language--action models such as RT-2, OpenVLA, and $\pi_{0.5}$ learn control from robot data \citep{rt2,openvla,pi05}. Code as Policies instead composes perception and control APIs, while Inner Monologue incorporates environmental feedback \citep{codepolicies,innermonologue}. CaP-X and Guava study tool, action, and observation design \citep{capx,guava}; Show-Harness and obstacle-aware harnesses address embodiment and geometric constraints \citep{showharness,safeharness}. RoboDawn couples discrete commands with feedback and demonstrations \citep{robodawn}. Thea and PhyAgentOS structure persistent state and outcome verification \citep{thea,phyagentos}, while Harness VLA and RoboHarness use execution memory to guide primitives or route heterogeneous policies \citep{harnessvla,roboharness}. These systems establish the importance of the interface around a model.

\paragraph{Self-improving agents.}
Reflexion retains verbal feedback, and GEPA evolves prompts from execution traces \citep{reflexion,gepa}. ADAS searches code-defined agents; SICA and DGM modify agent implementations \citep{adas,sica,dgm}. AutoHarness, Meta-Harness, and Self-Harness synthesize or revise executable harnesses from feedback \citep{autoharness,metaharness,selfharness}. Recursive Harness Self-Improvement evolves a prompt-level harness with a fixed optimizer \citep{rhi}, whereas STOP and Hyperagents also modify the improvement procedure \citep{stop,hyperagents}. RRSI regularizes harness edits, and Dream-RSI uses discovery histories to improve exploration \citep{rrsi,dreamrsi}. Hyperagents' robotics experiment evolves rewards for PPO controllers \citep[App.~C.3]{hyperagents}; PhysEvo instead uses the same frozen model for robot decisions and software revision.

\paragraph{Experience-driven embodied agents.}
REFLECT diagnoses failures from execution summaries \citep{reflect}; Voyager and RATs accumulate executable skills through interaction and autonomous play \citep{voyager,rats}. EmbodiSkill distinguishes faulty skills from failures to follow valid guidance \citep{embodiskill}; ASPIRE retains manipulation and debugging knowledge from multimodal traces \citep{aspire}. RoboRSI organizes skills hierarchically \citep{roborsi}, and Zetta evolves critics, recovery, and tools around a frozen action policy \citep{zetta}. SHAPER evolves skills and context-construction code with one frozen model serving as planner and optimizer, while fixing the action interface and executor \citep{shaper}. ENPIRE automates real-world policy development \citep{enpire}. RHO evolves multi-file robot repositories, including prompts, tools, and control code for LLM-in-the-loop execution \citep{rho}. HarnessPAI revises programs and repair skills across rollouts, keeping the selected high-level program fixed within each rollout \citep{harnesspai}. PhysEvo focuses on the joint evolution of skills, robot-facing action and observation interfaces, and diagnostic resources: the same frozen model acts through revised interfaces and reuses self-built tools to diagnose later failures.

\section{PhysEvo: Physical Recursive Self-Improvement}
\label{sec:method}
\begin{figure}[htbp]
\centering
\includegraphics[width=\linewidth]{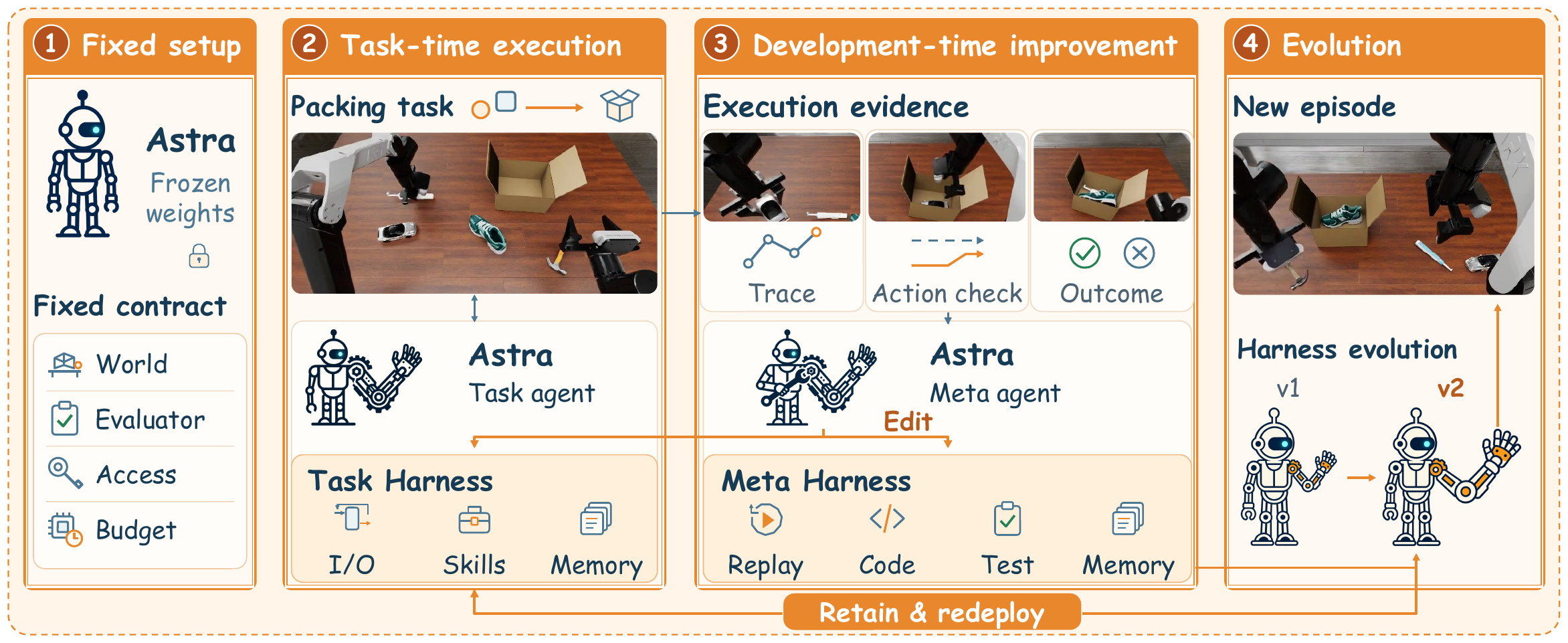}
\caption{Physical RSI in PhysEvo. A frozen Astra acts through the task harness and revises both task and meta resources from execution feedback. Accepted revisions support later episodes and improvement rounds.}
\label{fig:method}
\end{figure}

PhysEvo alternates between acting in the environment and revising the system used to act. A single frozen model performs both roles. Physical outcomes provide evidence; tested changes to the harness preserve what is learned from that evidence (\cref{fig:method}).

\subsection{One Frozen Model, Two Editable Harnesses}
Let $M_\theta$ denote Astra with fixed parameters. At iteration $r$, the persistent system consists of a task harness $\taskH_r$ and a meta harness $\metaH_r$:
\begin{equation}
H_r = (\taskH_r,\metaH_r).
\label{eq:harness}
\end{equation}
The task harness supplies observations, action tools, execution feedback, and skills. Task-agnostic skills guide operations such as grasp verification and recovery; task-specific skills guide activities such as pouring or packing. The meta harness provides code inspection and editing, trajectory review, diagnostic experiments, and revision testing. The meta-agent may modify either harness, including its own tools. The two roles use separate contexts and tool access but the same model. Model weights, robot dynamics, and the task evaluator are not editable.

Initialization provides a working observation--action loop with TCP targets and local IK, plus a general code-editing workspace. From this starting point, the system develops joint-level control, specialized observation tools, and reusable skills through execution and revision.

\subsection{Task Execution and Physical Feedback}
The task agent receives the instruction, selected skills, camera images, and robot proprioception. It selects an action or observation tool, then uses the returned result to make its next decision. Observation tools can retrieve evidence or calculate geometry without advancing the environment. Motion tools execute commands and return the resulting state and images.

For diagnosis, the harness links each requested action to its dispatched command and observed outcome. An unsuccessful grasp can thus be examined through approach, gripper closure, and subsequent object motion, rather than reduced to a final score. During execution, the same feedback supports closed-loop correction within the episode.

\subsection{Revision, Testing, and Recursive Reuse}
The meta-agent reviews execution alongside the current implementation, forms a failure hypothesis, and tests it through further inspection or diagnostic experiments. A revision may change action or observation tools, execution logic, skills, or diagnostic utilities. When a tool changes, its skill guidance is revised with it so that the agent inherits both a capability and instructions for using it.

Candidate revisions are tested during development, not on the held-out layouts used for final scoring. The meta-agent first checks that edited code runs and tools return the intended outputs. For behavioral changes, it compares the candidate with the current harness on the same task and initial scene. A score gain, or progress beyond the failure without a score decrease, supports adoption; a score decrease leads to rejection. A diagnostic-tool revision is checked against the error or measurement it is intended to address.

Accepted changes take effect between episodes. Changes to the task harness shape later execution; changes to the meta harness shape later diagnosis and revision. A review tool built in one round can therefore support a different correction in a later round. The improvement process thus inherits changes to its own machinery. A retained deployment version is then fixed for that task's held-out evaluation (\cref{sec:setup}). \Cref{app:algorithm} gives the full procedure.

\section{The Evolved Embodied Framework}
\label{sec:framework}
Physical RSI reshapes what Astra can control, what evidence it can obtain, and what guidance persists. We describe the resulting architecture below; tasks retain their own deployment versions (\cref{sec:setup}).

\subsection{Joint-Level Control}
The default \tool{control\_arms} interface accepts twelve absolute arm-joint targets and two continuous gripper commands. It supports both arms and remains available while an object is held. The executor enforces native actuator limits and checks measured motion; Astra chooses the approach, arm configuration, and reconfiguration.

TCP control asks the model for an end-effector goal and delegates posture selection to IK. Joint control instead makes posture part of the policy decision. This additional authority lets Astra adjust an obstructive configuration and continue manipulation, rather than repeatedly requesting a target from an unsuitable posture. \Cref{app:control} compares the interfaces under fixed model, skill, initialization, and budget.

\subsection{Active and Retrospective Observation}
Astra can move a free arm to obtain a useful wrist-camera view while the other arm manipulates an object. When the relevant event is no longer visible, \tool{review\_frames} retrieves earlier images and aligned robot states from the current episode without advancing physics. Active viewing seeks new evidence; historical retrieval recovers evidence already observed.

Pixel projection and multi-view triangulation compute geometry for image features selected by Astra. Temporal triangulation uses views collected as a wrist camera moves. The model decides what to inspect and where to look; deterministic tools compute geometry and check validity using images and known camera and robot geometry, not hidden object coordinates. Temporal reconstruction requires a stationary point and a translational viewing baseline.

\subsection{Reusable Skills and Diagnostic Tools}
Skills link tools to the physical conditions under which they are useful: when to verify a grasp, obtain another view, reorient an object, clear an obstacle, or check a release. The pouring skill, for example, maintains a free-arm side view through pouring and upright recovery. A packing skill requires wrist and gripper clearance above raised box flaps before turning. Skills retain these lessons while leaving action decisions to Astra's closed-loop policy.

The meta harness retains trajectory-review tools that align images, robot states, and policy decisions. The task agent uses runtime tools and selected skills; the meta-agent uses the broader editing and diagnostic workspace. Experience is therefore retained in two forms: guidance for acting and tools for understanding why an action failed.

\section{Experiments}
\label{sec:results}
We first evaluate retained deployment versions on held-out simulation layouts, then examine the revisions behind the resulting framework. On real hardware, we continue skill adaptation after deployment.

\subsection{Experimental Setup}
\label{sec:setup}
\paragraph{Tasks and metrics.}
We evaluate all 42 base RoboDojo simulation tasks across generalization, precision, long-horizon execution, memory, and open-ended instruction following. Native score ranges from 0 to 100; success rate (SR) counts scores of 100. Each task averages one attempt on each of five held-out layouts. The headline comparison weights the five dimensions equally after averaging tasks within each dimension. We also report the equal-task average and two task-level comparisons covering 16 distinct tasks (\cref{tab:galaxy10,tab:f2eight}).

\paragraph{Development, sharing, and scoring.}
Simulation self-improvement precedes scoring. Development resources are shared across tasks; at the task-agent level, only tools and task-agnostic skills are shared, while task-specific skills remain task-local. Each task retains a deployment version that stays fixed across its five scored layouts. Scored layouts are excluded from revision and version selection. Before each scored episode, we reset session context, historical-frame indices, and writable memory. Online correction and same-episode history are allowed, but persistent tools and skills do not change. \Cref{app:evaluation} details these boundaries.

\paragraph{Comparison methods.}
References include RoboDojo robot and language-model policies \citep{robodojo_board_20260923,zhang2026gpt6astrarobodojo}, GPT-as-Policy's direct Astra and Astra--$\pi_{0.5}$ hybrid \citep{su2026astra}, and RoboDawn's one-shot Astra agent with closed-loop feedback and in-context demonstrations \citep{robodawn}. Bold and underlining indicate the best and second-best values within each comparison panel.

\subsection{Performance Across RoboDojo}
Under five-dimension weighting, PhysEvo achieves \textbf{\DimensionScore{}/100 score and \DimensionSR{}\% SR}, exceeding RoboDawn one-shot's 47.17\% SR by 14.83 points (\cref{tab:dimensions}). This is the highest reported average among the compared methods, including an Astra agent already equipped with feedback and demonstrations. Equal weighting of all 42 tasks instead gives \AllTaskScore{} score and \AllTaskSR{}\% SR.

\Needspace{5\baselineskip}
Relative to RoboDawn, PhysEvo improves SR by 46.67 percentage points in memory, 17.50 in open-ended tasks, and 15.00 in precision. These gains combine stronger memory and instruction-dependent performance with improved precision (\cref{tab:dimensions}).

\begin{table}[!htbp]
\PhysTableBefore
\centering
\caption{RoboDojo performance across five capability dimensions. The average weights dimensions equally.}
\label{tab:dimensions}
\begingroup
\PhysTableSetup
\begin{tabularx}{\linewidth}{@{}>{\raggedright\arraybackslash}p{.18\linewidth}*{12}{R}@{}}
\toprule[0.8pt]
\textbf{Method} & \multicolumn{2}{c}{\textbf{General.}} & \multicolumn{2}{c}{\textbf{Precision}} & \multicolumn{2}{c}{\textbf{Long-horizon}} & \multicolumn{2}{c}{\textbf{Memory}} & \multicolumn{2}{c}{\textbf{Open}} & \multicolumn{2}{c}{\textbf{Average}} \\
 & \textit{Score} & \textit{SR (\%)} & \textit{Score} & \textit{SR (\%)} & \textit{Score} & \textit{SR (\%)} & \textit{Score} & \textit{SR (\%)} & \textit{Score} & \textit{SR (\%)} & \textit{Score} & \textit{SR (\%)} \\
\midrule
\multicolumn{13}{l}{\itshape Robot policies} \\
\midrule
Simate-beta & 35.09 & 27.95 & 34.35 & 26.92 & \underline{57.84} & \underline{43.42} & 33.33 & 33.00 & 9.12 & 8.50 & 33.95 & 27.96\\
Liber-0 Preview & 24.99 & 18.61 & \underline{38.28} & \underline{33.17} & 45.98 & 33.17 & 37.77 & 37.33 & 6.68 & 5.33 & 30.74 & 25.52\\
Liber-0 Lite & 25.36 & 19.11 & 36.60 & 31.08 & 45.50 & 32.92 & 35.68 & 35.44 & 3.06 & 2.58 & 29.24 & 24.23\\
DM0.5 & 15.78 & 10.95 & 24.82 & 16.75 & 33.70 & 19.50 & 47.74 & 47.44 & 2.43 & 2.08 & 24.90 & 19.34\\
GalaxeaVLA G0.5 & 18.46 & 12.83 & 28.25 & 20.42 & 44.12 & 32.25 & 8.61 & 7.33 & 1.73 & 1.58 & 20.23 & 14.88\\
Xiaomi-Robotics-1 & 23.55 & 17.00 & 26.69 & 18.83 & 38.39 & 23.67 & 7.81 & 6.56 & 3.94 & 3.58 & 20.07 & 13.93\\
OpenWAM-$\alpha$ & 20.71 & 14.84 & 18.45 & 9.25 & 34.93 & 25.33 & 10.41 & 9.11 & 1.41 & 1.08 & 17.18 & 11.92\\
$\pi_{0.5}$ & 13.38 & 8.17 & 12.40 & 5.50 & 23.54 & 14.67 & 5.89 & 4.67 & 1.98 & 1.67 & 11.44 & 6.93\\
\midrule
\multicolumn{13}{l}{\itshape Language-model policies} \\
\midrule
Official Astra & 33.36 & 30.50 & 12.65 & 4.00 & 21.45 & 8.25 & 43.04 & 38.67 & 34.36 & 31.00 & 28.97 & 22.48\\
GPT-5.5 & 0.20 & 0.00 & 0.42 & 0.00 & 0.51 & 0.00 & 1.67 & 1.67 & 2.87 & 2.75 & 1.13 & 0.88\\
DeepSeek-Flash & 2.42 & 1.67 & 1.75 & 0.00 & 2.12 & 0.00 & 2.17 & 1.67 & 6.50 & 6.25 & 2.99 & 1.92\\
RoboDawn (1-shot) & \textbf{55.08} & \textbf{50.00} & 35.88 & 25.00 & \textbf{59.88} & \textbf{45.00} & \underline{54.33} & \underline{53.33} & \underline{68.00} & \underline{62.50} & \underline{54.63} & \underline{47.17}\\
\rowcolor{black!8}
\textbf{PhysEvo (ours)} & \underline{51.08} & \underline{45.00} & \textbf{50.38} & \textbf{40.00} & 56.38 & \textbf{45.00} & \textbf{100.00} & \textbf{100.00} & \textbf{82.88} & \textbf{80.00} & \textbf{68.14} & \textbf{62.00}\\
\bottomrule[0.8pt]
\end{tabularx}
\PhysNote{RoboDawn uses GPT-6 Astra with one in-context demonstration; its Score and SR follow the public evaluation records.}
\endgroup

\PhysTableAfter
\end{table}

\FloatBarrier
\subsection{Comparison with Direct and Hybrid Astra Policies}
On the ten-task GPT-as-Policy set, PhysEvo reaches \textbf{\TenTaskSR{}\% SR}, versus 26.00\% for direct Astra and 48.00\% for the hybrid (\cref{tab:galaxy10}). The three systems organize Astra's capabilities differently: direct action, delegation to a trained VLA, or action through an experience-revised harness.

PhysEvo improves SR over Direct on eight tasks and ties on two. Gains span language-conditioned classification, sorting imitation, and number arrangement, as well as packing and bottle disposal. An experience-revised harness thus offers a route to higher aggregate performance without a separately trained action policy.

\begin{table}[!htbp]
\PhysTableBefore
\centering
\caption{Comparison on the ten-task GPT-as-Policy evaluation set. Hybrid combines Astra with $\pi_{0.5}$; Direct uses Astra alone.}
\label{tab:galaxy10}
\begingroup
\PhysTableSetup
\begin{tabularx}{\linewidth}{@{}>{\raggedright\arraybackslash}p{.20\linewidth}*{12}{R}@{}}
\toprule[0.8pt]
\textbf{Task} & \multicolumn{2}{c}{G0.5} & \multicolumn{2}{c}{Xiaomi} & \multicolumn{2}{c}{O-WAM} & \multicolumn{2}{c}{Hybrid} & \multicolumn{2}{c}{Direct} & \multicolumn{2}{c}{\textbf{PhysEvo}} \\
 & \textit{Score} & \textit{SR} & \textit{Score} & \textit{SR} & \textit{Score} & \textit{SR} & \textit{Score} & \textit{SR} & \textit{Score} & \textit{SR} & \textit{Score} & \textit{SR} \\
\midrule
Organize table & 46.33 & 11.33 & 57.67 & \textbf{14.67} & \textbf{62.50} & \underline{13.33} & \underline{60.00} & 0.00 & 30.00 & 0.00 & \underline{60.00} & 0.00\\
Classify by language & 1.07 & 0.00 & 2.00 & 0.67 & 1.33 & 0.00 & 38.00 & 20.00 & \underline{60.00} & \underline{40.00} & \textbf{100.00} & \textbf{100.00}\\
Imitate sorting & 1.67 & 0.00 & 2.50 & 0.67 & 2.90 & 0.67 & \underline{53.00} & \underline{40.00} & 0.00 & 0.00 & \textbf{100.00} & \textbf{100.00}\\
Largest number & 4.11 & 0.53 & 8.56 & 3.73 & 4.36 & 0.53 & 50.00 & \underline{40.00} & \underline{57.00} & \underline{40.00} & \textbf{100.00} & \textbf{100.00}\\
Pack objects & 17.12 & 2.93 & 18.69 & 2.40 & 20.83 & 3.73 & \underline{50.00} & \underline{20.00} & \underline{50.00} & \underline{20.00} & \textbf{70.00} & \textbf{40.00}\\
Classify objects & 10.33 & 4.00 & 17.20 & 8.67 & 5.53 & 1.33 & \underline{71.00} & \underline{60.00} & \textbf{100.00} & \textbf{100.00} & \textbf{100.00} & \textbf{100.00}\\
Build tower & \textbf{82.93} & \textbf{78.67} & 52.60 & 47.33 & 52.53 & 38.67 & \underline{64.00} & \underline{60.00} & 12.00 & 0.00 & 40.00 & 20.00\\
Make kong & \textbf{90.00} & \textbf{90.00} & 41.33 & 41.33 & 32.00 & 32.00 & 40.00 & 40.00 & 0.00 & 0.00 & \underline{60.00} & \underline{60.00}\\
Fold clothes & 32.75 & 28.00 & 41.49 & 38.93 & 51.31 & 49.07 & \textbf{100.00} & \textbf{100.00} & 40.00 & 40.00 & \underline{68.00} & \underline{60.00}\\
Bottles into dustbin & 96.30 & 94.00 & \underline{97.70} & \underline{96.67} & 94.03 & 91.33 & \textbf{100.00} & \textbf{100.00} & 36.00 & 20.00 & \textbf{100.00} & \textbf{100.00}\\
\midrule
\textbf{Mean} & 38.26 & 30.95 & 33.97 & 25.51 & 32.73 & 23.07 & \underline{62.60} & \underline{48.00} & 38.50 & 26.00 & \textbf{79.80} & \textbf{68.00}\\
\bottomrule[0.8pt]
\end{tabularx}
\PhysNote{Score is on a 0--100 scale; SR is in percent. The three VLA references have the highest ten-task average scores in the source report \citep{su2026astra}. G0.5 = GalaxeaVLA (G0.5); Xiaomi = Xiaomi-Robotics-1; O-WAM = OpenWAM-$\alpha$.}
\endgroup

\PhysTableAfter
\end{table}

\FloatBarrier
\subsection{Manipulation Tasks Challenging Direct Astra}
The RoboDojo Astra report identifies eight tasks with VLA SR $\geq20\%$ and direct-Astra SR $<5\%$ \citep{zhang2026gpt6astrarobodojo}. This externally defined subset tests whether improvement reaches manipulation tasks that direct Astra rarely completes.

PhysEvo achieves \textbf{\EightTaskSR{}\% SR and \EightTaskScore{} score}, compared with official Astra's 1.25\% and 5.96: a 53.75-point SR gain (\cref{tab:f2eight}). Its mean SR also exceeds Simate-beta's 50.83\%, Liber-0 Preview's 46.44\%, and RoboDawn's 45.00\%.

On the two pouring tasks, PhysEvo improves SR over RoboDawn from 0\% to 80\% for pouring balls and from 80\% to 100\% for pouring liquid. These gains align with the observation and pouring cases below.

\begin{table}[!htbp]
\PhysTableBefore
\centering
\caption{Eight manipulation tasks challenging direct Astra. PhysEvo results average five layouts per task. The task set is defined by the RoboDojo report, with VLA SR $\geq20\%$ and Astra SR $<5\%$.}
\label{tab:f2eight}
\begingroup
\PhysTableSetup
\begin{tabularx}{\linewidth}{@{}>{\raggedright\arraybackslash}p{.20\linewidth}*{12}{R}@{}}
\toprule[0.8pt]
\textbf{Task} & \multicolumn{2}{c}{Official Astra} & \multicolumn{2}{c}{G0.5} & \multicolumn{2}{c}{Simate-beta} & \multicolumn{2}{c}{Liber-0 Preview} & \multicolumn{2}{c}{RoboDawn} & \multicolumn{2}{c}{\textbf{PhysEvo}} \\
 & \textit{Score} & \textit{SR} & \textit{Score} & \textit{SR} & \textit{Score} & \textit{SR} & \textit{Score} & \textit{SR} & \textit{Score} & \textit{SR} & \textit{Score} & \textit{SR} \\
\midrule
Make kong & 0.00 & 0.00 & \textbf{90.00} & \textbf{90.00} & \underline{76.00} & \underline{76.00} & 18.00 & 18.00 & 60.00 & 60.00 & 60.00 & 60.00\\
Build tower & 16.40 & 2.00 & 82.93 & 78.67 & \textbf{87.40} & \textbf{83.33} & \underline{84.33} & \underline{81.00} & 54.00 & 40.00 & 40.00 & 20.00\\
Insert tubes & 6.00 & 0.00 & 58.53 & 42.67 & 41.20 & 22.00 & \underline{82.53} & \underline{73.00} & \textbf{88.00} & \textbf{80.00} & \textbf{88.00} & \textbf{80.00}\\
Tic-tac-toe & 10.70 & 0.00 & 65.23 & 40.00 & 86.27 & 60.67 & \underline{95.67} & \underline{89.00} & \textbf{100.00} & \textbf{100.00} & \textbf{100.00} & \textbf{100.00}\\
Pour balls into vase & 4.00 & 4.00 & 28.00 & 28.00 & \underline{45.33} & \underline{45.33} & 30.67 & 31.00 & 0.00 & 0.00 & \textbf{80.00} & \textbf{80.00}\\
Pour liquid into cup & 4.00 & 4.00 & 35.34 & 35.34 & 55.33 & 55.33 & 34.67 & 34.50 & \underline{80.00} & \underline{80.00} & \textbf{100.00} & \textbf{100.00}\\
Laptop / headphones & 1.60 & 0.00 & 28.27 & 11.34 & \textbf{53.87} & \textbf{34.00} & \underline{38.14} & \underline{20.00} & 12.00 & 0.00 & 16.00 & 0.00\\
Fill pen holder & 5.00 & 0.00 & 41.27 & 18.67 & \textbf{50.37} & \textbf{30.00} & \underline{49.77} & \underline{25.00} & 26.00 & 0.00 & 19.00 & 0.00\\
\midrule
\textbf{Mean} & 5.96 & 1.25 & 53.70 & 43.09 & \underline{61.97} & \underline{50.83} & 54.22 & 46.44 & 52.50 & 45.00 & \textbf{62.88} & \textbf{55.00}\\
\bottomrule[0.8pt]
\end{tabularx}
\PhysNote{SR is in percent. G0.5 = GalaxeaVLA (G0.5); RoboDawn uses GPT-6 Astra, 1-shot. For Simate-beta and Liber-0 Preview, Std/Rand Score and SR are averaged within task before the eight-task mean. Further references appear in \cref{app:f2references}.}
\endgroup

\PhysTableAfter
\end{table}

\FloatBarrier
\subsection{What Changes Through Physical RSI?}
\label{sec:cases}
Three cases trace execution feedback to concrete changes in control, observation, and diagnosis.

\begin{figure}[H]
\centering
\includegraphics[width=\linewidth]{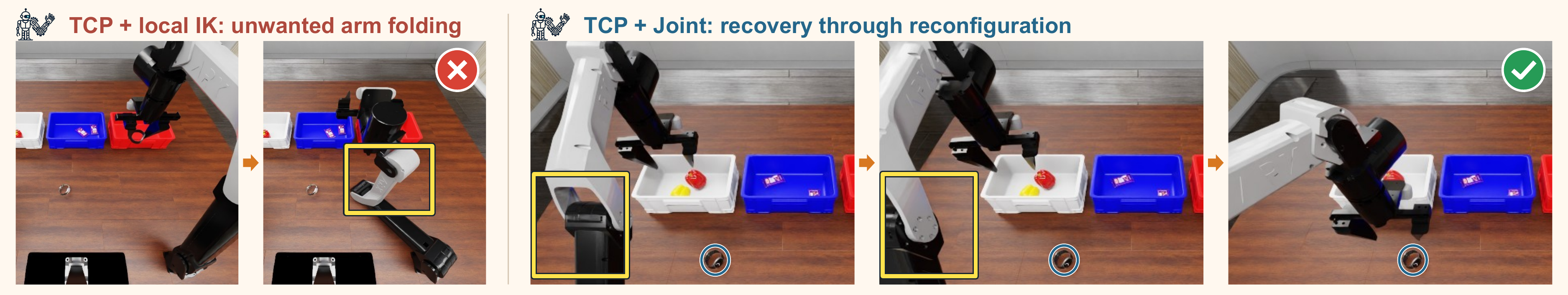}
\caption{Joint reconfiguration enables recovery from an obstructive arm posture.}
\label{fig:case_control}
\end{figure}

\Needspace{8\baselineskip}
\paragraph{Making posture a policy decision.}
\Cref{fig:case_control} shows two independent executions: an obstructive fold under TCP control with local IK, and recovery from a rejected TCP motion through joint reconfiguration. End-effector goals leave posture to IK; joint control makes that missing choice expressible. In a focused two-layout comparison with the model, skill, initialization, and budget fixed, mean scores are 40 for TCP, 100 for Joint, and 55 for Both (six episodes; \cref{app:control}). Joint outperforming Both in this comparison also shows that a larger tool menu alone is not sufficient.

\begin{figure}[H]
\centering
\includegraphics[width=\linewidth]{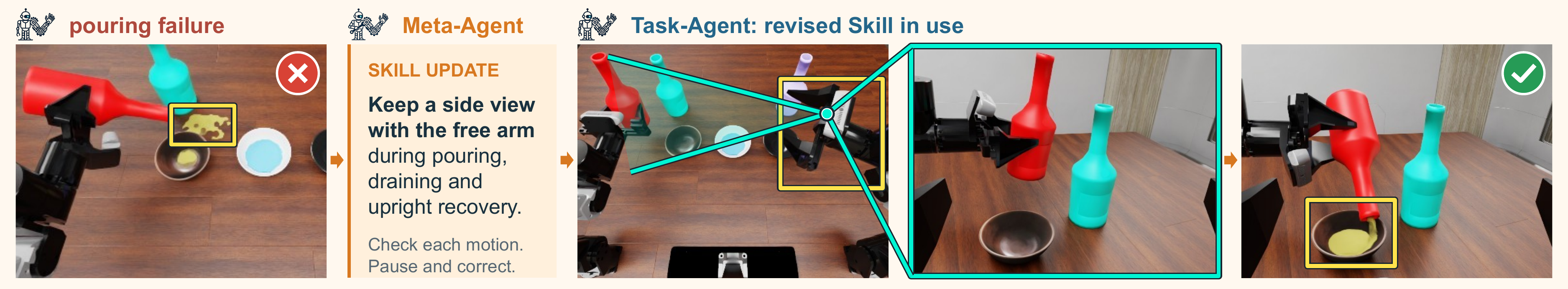}
\caption{Skill revision maintains a useful view throughout pouring and recovery.}
\label{fig:case_observation}
\end{figure}

\Needspace{8\baselineskip}
\paragraph{Maintaining evidence during action.}
A pouring failure occurs while the bottle blocks the holding arm's wrist view. Revising the pouring skill, with the other loaded skills fixed, directs the free arm to maintain a side view through pouring and recovery. The task agent then checks the bottle mouth, cup rim, liquid landing point, and clearance, corrects its motion, and completes the task (\cref{fig:case_observation}). The revised skill keeps existing sensors useful as manipulation alters visibility.

\begin{figure}[H]
\centering
\includegraphics[width=\linewidth]{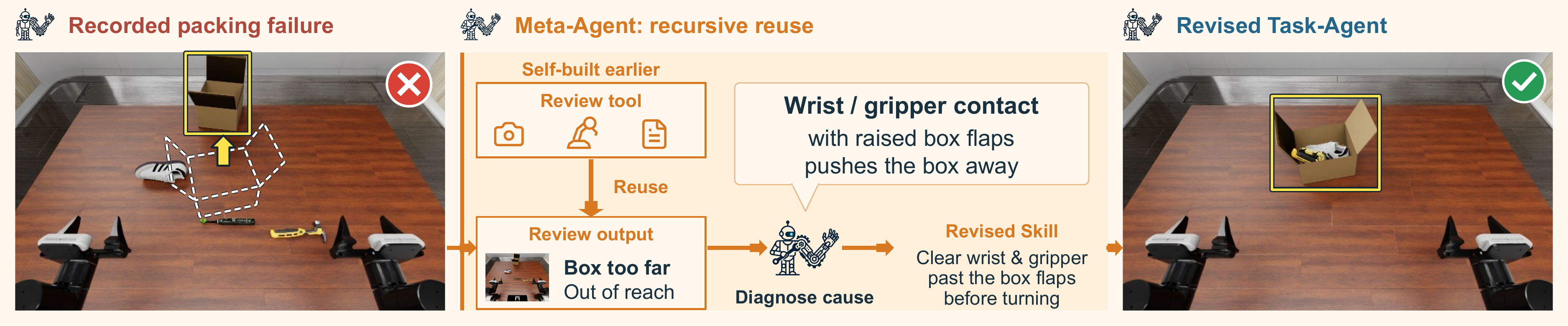}
\caption{An inherited review tool supports a later packing-skill revision.}
\label{fig:case_diagnosis}
\end{figure}

\Needspace{8\baselineskip}
\paragraph{Inheriting a means of diagnosis.}
In \cref{fig:case_diagnosis}, the meta-agent invokes a review tool built in an earlier round. Aligned images, robot states, and policy decisions help trace box displacement to wrist or gripper contact with raised flaps. The skill is then revised to require clearance above the flaps before turning. In the revised execution, the task agent follows this guidance and avoids pushing the box away. The final symptom, a box out of reach, thus leads to a specific intervention on the preceding contact. Crucially, a diagnostic resource created by one improvement round helps produce a correction in another: the system inherits a way to improve, not only a way to act.

\FloatBarrier
\subsection{Real-World Adaptation through Skill Revision}
\label{sec:hardware}
\begin{figure}[tbp]
\centering
\includegraphics[width=0.9\linewidth]{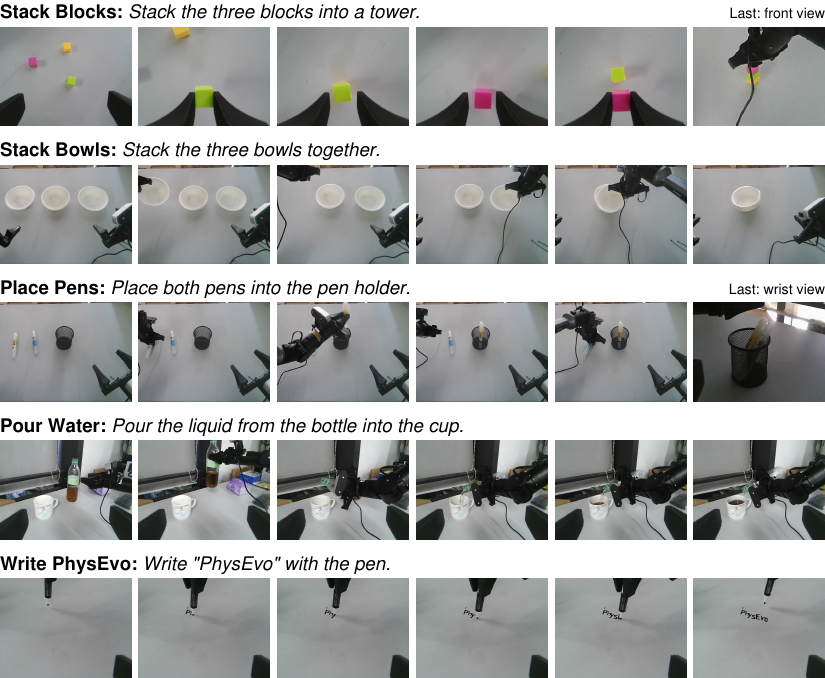}
\caption{Five AgileX PiPER tasks after trajectory-driven skill revision. Each row shows one execution in time; alternate final views are marked.}
\label{fig:hardware}
\end{figure}

We deploy the simulation-evolved harness on an AgileX PiPER platform and continue the execution--revision loop. After each attempt, the meta-agent analyzes the trajectory and revises skills; the task agent uses those revisions in the next attempt. These cycles yield successful executions of block stacking, bowl stacking, pen placement, liquid pouring, and writing ``PhysEvo'' with a pen (\cref{fig:hardware}). \paragraph{Hardware results.}
We report five trials per task, for 25 trials in total (\cref{tab:hardware_results}). PhysEvo achieves a mean score of 90.60/100 and 84.00\% SR. Block stacking, bowl stacking, and pen placement each achieve 100\% SR; pouring and writing each achieve 60\% SR. \Cref{app:hardware_scoring} specifies the task-level scoring rules.

\begin{table}[H]
\centering
\caption{Real-world performance on AgileX PiPER. Each task has five trials.}
\label{tab:hardware_results}
\begingroup\normalsize
\setlength{\tabcolsep}{14pt}\renewcommand{\arraystretch}{1.12}
\begin{tabular}{@{}lrr@{}}
\toprule
Task & Score & SR (\%) \\
\midrule
Stack blocks & 100.00 & 100.00 \\
Stack bowls & 100.00 & 100.00 \\
Place pens & 100.00 & 100.00 \\
Pour water & 60.00 & 60.00 \\
Write PhysEvo & 93.00 & 60.00 \\
\midrule
\textbf{Overall} & \textbf{90.60} & \textbf{84.00} \\
\bottomrule
\end{tabular}
\endgroup
\end{table}

Two revisions illustrate how physical discrepancies change the inherited skills.

\paragraph{Compensating for liquid motion.}
The liquid stream's forward momentum causes it to overshoot the cup. The meta-agent revises the pouring skill to shift the bottle opening backward relative to the cup, compensating for the observed displacement. The hardware revision changes the action to account for the observed flow.

\paragraph{Changing when geometry is used.}
Our physical arm has lower positioning accuracy than its simulated counterpart, and localization from the initial distant view is often unreliable. The revised skill reverses the geometry-first sequence: visual feedback guides a coarse approach, then geometry provides close-range localization for fine positioning. By changing the order of perception and action, the revised skill uses geometry under more reliable viewing conditions.

These cases demonstrate continued post-deployment adaptation. Additional simulation executions appear in \cref{app:simulation}.

\section{Discussion and Conclusion}
PhysEvo uses physical consequences to revise the tools, skills, and diagnostic resources of a frozen Astra. Its retained deployment versions achieve \DimensionSR{}\% success under five-dimension weighting across 42 RoboDojo tasks. The cases identify three ways experience changes the system: exposing posture control, maintaining useful observations, and inheriting tools for later diagnosis. On AgileX PiPER, the same loop adapts skills to liquid motion and localization discrepancies, with a mean score of 90.60/100 and 84.00\% success across five tasks. The central lesson is that experience can improve not only the next action, but the interface and diagnostic resources available to future actions.

\FloatBarrier

\clearpage
\appendix
\section{Physical RSI Procedure}
\label{app:algorithm}
At iteration $r$, the persistent system is $H_r=(\taskH_r,\metaH_r)$. Both roles use the same frozen model $M_\theta$. The task agent selects tools from a context assembled by the task harness:
\begin{equation}
 a_t\sim M_\theta(\cdot\mid P_{\taskH_r}(\ell,o_{\leq t},m_t)),\qquad
 (o_{t+1},f_t)=\operatorname{RunTool}_{\taskH_r}(a_t).
\end{equation}
Here $\ell$ is the instruction, $o_{\leq t}$ contains observations, $m_t$ is episode memory, and $f_t$ is execution feedback. The index $t$ counts tool interactions. A motion call can execute several native steps; a review or geometry call need not advance the environment.

Let $E_r$ contain feedback available to the meta-agent. One improvement round is summarized by
\begin{align}
 \widetilde H_r &\leftarrow \operatorname{Revise}_{M_\theta,\metaH_r}(H_r,E_r),\\
 y_r &\leftarrow \operatorname{Test}(\widetilde H_r,H_r),\\
 H_{r+1} &\leftarrow \operatorname{Select}_{M_\theta,\metaH_r}(H_r,\widetilde H_r,y_r).
\end{align}
$\widetilde H_r$ is a candidate revision and $y_r$ contains its test results. The selection step either adopts the candidate or preserves the current harness. Robot dynamics, actuator restrictions, model configuration, and evaluation rules are outside the editable system.

\begin{table}[H]
\centering
\caption{Physical RSI procedure in PhysEvo.}
\begingroup\PhysTableSetup\fontsize{10}{13}\selectfont
\begin{tabularx}{\linewidth}{@{}rL@{}}
\toprule
 & \textbf{Inputs:} frozen Astra $M_\theta$, initial task and meta harnesses $H_0$, and an interaction budget.\\
\midrule
1 & Execute a task using the current task harness.\\
2 & Give the meta-agent the task outcome, relevant images and robot states, action feedback, and current code.\\
3 & Let the meta-agent inspect the failure, retrieve evidence, and run diagnostic experiments.\\
4 & Propose a candidate revision to tools, skills, execution logic, or meta resources.\\
5 & Run the edited code and check its outputs. For task-behavior changes, compare the candidate with the current harness on the same task and initial scene.\\
6 & Adopt a candidate with a score gain, or progress beyond the failure without a score decrease. Reject a candidate that lowers the score. Test diagnostic changes against the error or measurement they address.\\
7 & Load accepted tools and their associated skills together in subsequent episodes. Reuse accepted meta resources in subsequent improvement rounds.\\
8 & Repeat within the interaction budget, then retain a deployment version for that task and evaluate held-out layouts without further updates.\\
\bottomrule
\end{tabularx}
\endgroup
\end{table}

\paragraph{Initialization and persistence.}
The initial task harness supplies instructions, camera observations, proprioception, and a TCP/local-IK action interface. The initial meta harness supplies a code-editing and execution workspace. Retained tools and skills form a compatible system: an updated tool and the guidance for using its API are loaded together. Updates take effect between episodes rather than changing an active episode's tools. Selected skills are loaded at episode start. The task agent uses its runtime tools; the meta-agent uses the code-editing workspace. Development resources are shared across tasks. At the task-agent level, tools and task-agnostic skills are shared, while task-specific skills remain task-local. Each task retains its own deployment version before scoring.

\section{Evaluation Details}
\label{app:evaluation}
\subsection{Metrics and Aggregation}
For task $t$, let $s_{t\ell}$ be the native score on layout $\ell\in\{1,\ldots,5\}$. We compute
\begin{equation}
 M_t=\frac{1}{5}\sum_{\ell=1}^{5}s_{t\ell},\qquad
 \mathrm{SR}_t(\%)=\frac{100}{5}\sum_{\ell=1}^{5}\mathbf{1}[s_{t\ell}=100].
\end{equation}
The ten-task GPT-as-Policy set and the eight-task manipulation subset overlap on tower building and making kong. Their union contains 16 distinct tasks and 80 attempts, giving \FocalTaskScore{} score and \FocalTaskSR{}\% SR. Each task is counted once in this union; the two subset averages are not averaged together.

\paragraph{Development and held-out scoring.}
Self-improvement takes place during simulation development. For each task, a retained deployment version is fixed across its five scored layouts; these layouts are not used for harness revision or version selection. Development resources may carry experience across tasks, but task-agent sharing is restricted to tools and task-agnostic skills. Task-specific skills remain task-local.

\paragraph{Episode state and information access.}
Before every scored episode, session context, historical-frame indices, and writable memory are reset. The task agent loads its retained tools and selected skills, then may accumulate and revisit observations from that episode and correct its actions online. This within-episode feedback does not change persistent tools, skills, or the deployment version. Scored instances do not supply privileged information to the meta-agent or contribute to revision or version selection. The hardware experiments use a separate adaptation protocol with skill revision after deployment; \cref{sec:hardware} reports the real-world trial outcomes.

\subsection{Comparison Sources and Aggregation}
\paragraph{Five-dimension comparison.}
The robot and direct-language-model results in \cref{tab:dimensions} are taken from the RoboDojo leaderboard and Astra report \citep{robodojo_board_20260923,zhang2026gpt6astrarobodojo}. The final average weights the five dimensions equally. RoboDawn's five-dimension Score and SR are computed from its public evaluation records: GPT-6 Astra, 1-shot, with five runs per task \citep{robodawn_results}. The dimension-weighted means, 54.63 score and 47.17\% SR, agree with its paper \citep{robodawn}.

\paragraph{Direct and hybrid Astra policies.}
The ten-task set follows GPT-as-Policy \citep{su2026astra}. PhysEvo is evaluated on five layouts per task, with one attempt per layout, for 50 attempts in total. The source reports 50 selected cases for the hybrid and 48 valid scores for Direct. We recompute Direct's score by averaging task means, giving 38.50 rather than its pooled-case score of 37.81. The three VLA methods displayed in \cref{tab:galaxy10} have the highest ten-task average scores among the source report's VLA references.

\paragraph{Manipulation-challenge subset.}
The eight tasks in \cref{tab:f2eight} correspond to Table F2 of the RoboDojo Astra report \citep{zhang2026gpt6astrarobodojo}. Its selection criterion is VLA SR $\geq20\%$ and Astra SR $<5\%$. PhysEvo is evaluated on five scenes per task on this subset, with one attempt per scene. The main table uses the same six methods as \cref{fig:overview}. Simate-beta and Liber-0 Preview follow the 23 September 2026 leaderboard update, recorded on 26 September. RoboDawn uses GPT-6 Astra, 1-shot, with five runs per task \citep{robodawn_results}. Both Score and SR are shown in the task rows. \Cref{app:f2references} retains the other robot-policy references.

\Needspace{12\baselineskip}
\subsection{Real-World Scores and Success Criteria}
\label{app:hardware_scoring}
Each hardware task is evaluated over five trials. As in simulation, Score summarizes partial and full task completion on a 0--100 scale, while SR is the percentage of trials reaching the task's maximum score. Averaging the five task-level metrics in \cref{tab:hardware_results} gives a Score of 90.60/100 and an SR of 84.00\%. The task-specific scores are defined below.

\paragraph{Stack blocks.}
The goal is to stack three blocks. A score of 0 means that no block stacking is completed; 15 means that any two of the three blocks are stacked; 100 requires all three blocks to be stacked and the arm to return home.

\paragraph{Stack bowls.}
The goal is to stack three bowls. A score of 0 means that no bowls are correctly stacked; 15 means that any two bowls are stacked upright; 100 requires all three bowls to be stacked upright, the bottom bowl to rest stably on the table, and the arm to return home.

\paragraph{Place pens.}
The goal is to place both pens in the holder. Scores are 0 if neither pen is in the holder, 50 if either one is in the holder, and 100 if both are in the holder.

\paragraph{Pour water.}
The goal is to pour liquid from the bottle into the cup. A score of 0 means that the liquid has not been successfully poured into the cup. A score of 100 requires the cup to contain the required liquid and the bottle to be upright.

\paragraph{Write PhysEvo.}
The goal is to write ``PhysEvo'' on the table. The score is 0 when no letter is correct, and 15, 30, 45, 60, 75, or 90 when any one, two, three, four, five, or six letters are correct, respectively. All seven letters correct earns 100. The five scores are 100, 100, 100, 75, and 90, with a mean Score of 93/100 and an SR of 60\%.

\Needspace{12\baselineskip}
\section{Additional Framework Analysis}
\label{app:analysis}
\subsection{Control-Interface Comparison}
\label{app:control}
A focused comparison on language-based object classification evaluates TCP, Joint, and Both interfaces on two scene layouts, for six episodes in total. The model, task skill, initialization for each layout, and budget are held fixed. Mean scores are 40 for TCP, 100 for Joint, and 55 for Both. Joint is retained as the default control interface.

The control sequences in \cref{fig:case_control} illustrate two independent executions. The first develops an obstructive posture under TCP/local IK. In the second, Astra adjusts joints after a TCP rejection and then completes a motion to the same target pose. The mechanism is direct access to the arm configuration, which an end-effector target alone leaves to the IK solver.

\subsection{Observation Tools}
Historical-frame retrieval returns images and aligned robot states from earlier in the same episode. It does not advance the environment. Active viewing can place the free wrist camera to the side of a manipulation. Geometric tools combine model-selected image features with camera calibration and robot geometry.

A synthetic-marker reconstruction test accepted 8 of 12 points, with a median error of 0.615 mm and a 95th-percentile error of 1.437 mm. Four points were rejected by the validity checks. A separate wrist-camera translation of approximately 57.525 mm yielded a marker error of about 2.14 mm. These component tests measure the calibration and reconstruction paths. Temporal triangulation additionally requires correspondence to the same stationary point across views.

\subsection{Skill-Guided Observation and Diagnosis}
The pouring example changes the skill guidance while keeping the other loaded skills fixed. The revised skill maintains a free-arm side view during pouring, draining, and upright recovery. The task agent uses the observation arm while the pouring arm makes corrections, and the revised episode reaches task success.

The packing example reuses a trajectory-review tool that presents RGB, robot states, and policy decisions. The meta-agent inspects the box displacement and revises the packing skill to require clearance above the flaps before a turn. With the revised skill, the task agent avoids pushing the box away. The reused diagnostic tool thus supports a subsequent skill revision.

\clearpage
\section{Robot and Agent Averages on the Manipulation Subset}
\label{app:f2references}
The original 40 robot-policy means are reaggregated from the RoboDojo Astra report's task matrix \citep{zhang2026gpt6astrarobodojo}, weighting all eight tasks equally. The Simate-beta, Liber-0 Preview, and RoboDawn rows summarize the task-level Score and SR entries in \cref{tab:f2eight} \citep{robodojo_board_20260923,robodawn_results}. Each method is evaluated on the same task set and averaged with equal task weights. G0.5 is the strongest robot policy in the original 40-method set (43.09\% SR); Simate-beta is the strongest displayed robot-policy entry in the extended comparison (50.83\%). PhysEvo reaches \EightTaskSR{}\%.

\begin{table}[H]
\centering
\caption{Eight-task means for the original references and the extended comparison. Bold and underlining mark the best and second-best values.}
\label{tab:f2means}
\begin{minipage}[t]{0.485\linewidth}
\vspace{0pt}
\centering
\begingroup
\fontsize{10}{13}\selectfont
\setlength{\tabcolsep}{3pt}
\renewcommand{\arraystretch}{1.13}
\begin{tabularx}{\linewidth}{@{}Lrr@{}}
\toprule
\textbf{Method} & \textbf{Mean score} & \textbf{SR (\%)} \\
\midrule
\rowcolor{black!8}
\textbf{PhysEvo (ours)} & \textbf{62.88} & \textbf{55.00}\\
Simate-beta & \underline{61.97} & \underline{50.83}\\
Liber-0 Preview & 54.22 & 46.44\\
RoboDawn (1-shot) & 52.50 & 45.00\\
GalaxeaVLA (G0.5) & 53.70 & 43.09\\
Xiaomi-Robotics-1 & 44.73 & 31.83\\
DM0.5 & 37.80 & 26.42\\
OpenWAM-$\alpha$ & 34.10 & 26.25\\
VLAct & 28.26 & 22.67\\
Meituan-Robotics-0 & 26.76 & 16.25\\
StarVLA-PI\_v3 & 24.38 & 18.50\\
InternVLA-A1.5 & 23.84 & 16.00\\
Spatial Forcing & 23.78 & 16.42\\
Hy-Embodied-0.5-VLA & 23.61 & 15.17\\
StarVLA-OFT & 22.68 & 15.08\\
X-VLA & 22.12 & 15.25\\
Pi-05 & 19.73 & 12.00\\
StarVLA-GR00T & 15.95 & 9.00\\
StarVLA & 13.84 & 8.33\\
Xiaomi-Robotics-0 & 12.93 & 8.33\\
MolmoAct2 & 11.62 & 5.17\\
EventVLA & 11.42 & 5.92\\
GalaxeaVLA (G0) & 10.48 & 5.25\\
\bottomrule
\end{tabularx}
\endgroup
\end{minipage}
\hfill
\begin{minipage}[t]{0.485\linewidth}
\vspace{0pt}
\centering
\begingroup
\fontsize{10}{13}\selectfont
\setlength{\tabcolsep}{3pt}
\renewcommand{\arraystretch}{1.13}
\begin{tabularx}{\linewidth}{@{}Lrr@{}}
\toprule
\textbf{Method} & \textbf{Mean score} & \textbf{SR (\%)} \\
\midrule
GigaWorld-Policy-0 & 7.87 & 3.09\\
LingBot-VLA & 7.49 & 3.58\\
AHA-WAM & 7.29 & 2.50\\
X-WAM & 7.11 & 2.17\\
Pi-0 & 5.31 & 2.42\\
ABot-M0 & 4.61 & 1.67\\
SmolVLA (Single Task) & 3.22 & 1.17\\
GROOT-N1.7 & 3.22 & 1.25\\
InternVLA-A1 & 2.96 & 1.33\\
LDA-1B & 2.64 & 0.58\\
ACT (Single Task) & 2.58 & 1.75\\
GO-1 & 2.35 & 1.25\\
Fast-WAM & 1.58 & 0.00\\
RDT & 0.60 & 0.08\\
Dexora-1B & 0.48 & 0.00\\
H-RDT & 0.42 & 0.00\\
DM0 & 0.31 & 0.08\\
A1 & 0.11 & 0.00\\
TinyVLA & 0.09 & 0.00\\
OpenVLA-OFT & 0.09 & 0.00\\
Spirit v1.5 & 0.07 & 0.00\\
\midrule
Official Astra & 5.96 & 1.25\\
\bottomrule
\end{tabularx}
\endgroup
\end{minipage}
\end{table}

\clearpage
\section{Additional Simulation Executions}
\label{app:simulation}
\begin{figure}[H]
\centering
\includegraphics[width=\linewidth]{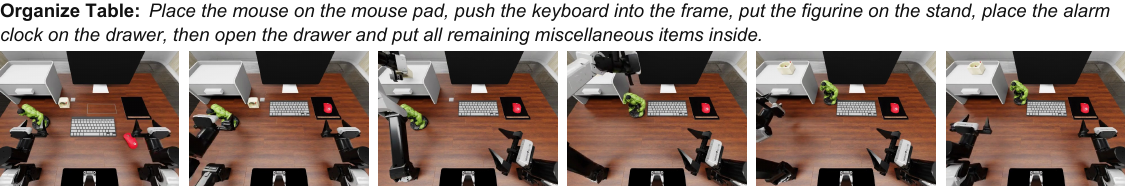}
\par\vspace{6pt}
\includegraphics[width=\linewidth]{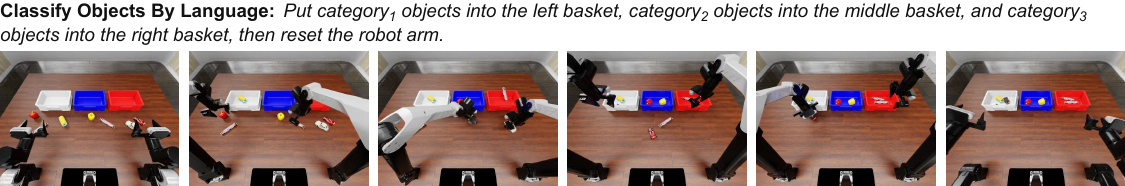}
\par\vspace{6pt}
\includegraphics[width=\linewidth]{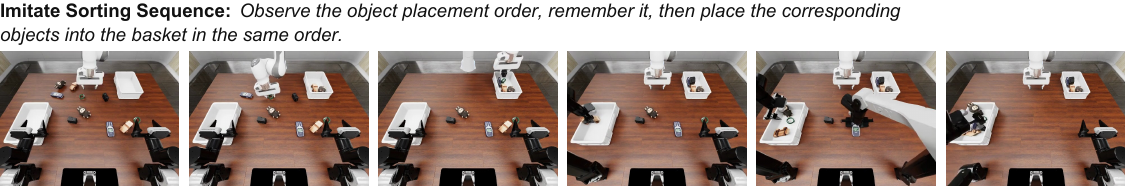}
\par\vspace{6pt}
\includegraphics[width=\linewidth]{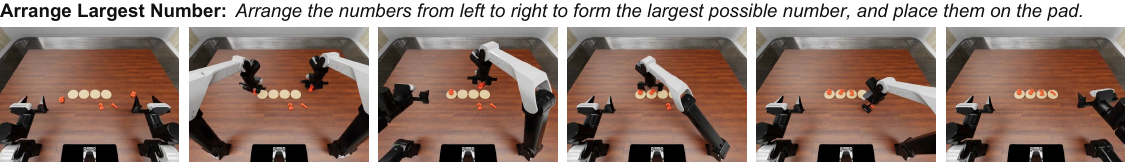}
\par\vspace{6pt}
\includegraphics[width=\linewidth]{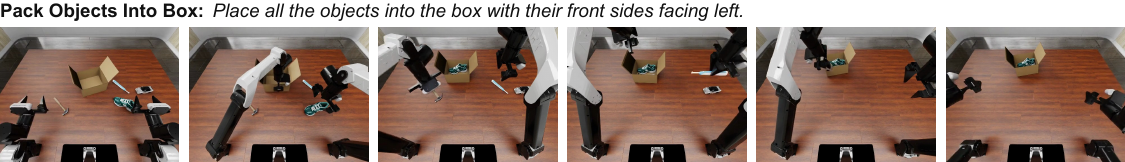}
\par\vspace{6pt}
\includegraphics[width=\linewidth]{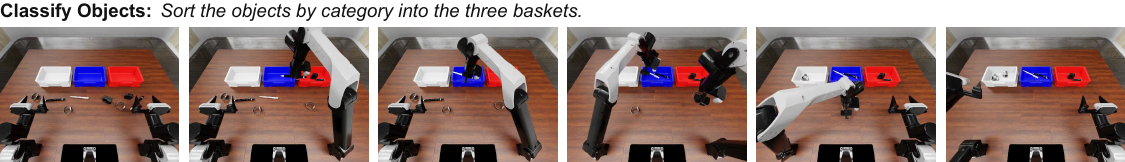}
\par\vspace{6pt}
\includegraphics[width=\linewidth]{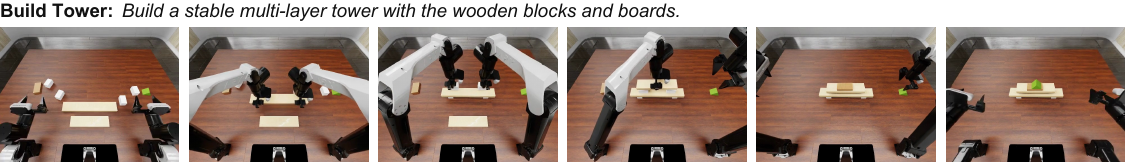}
\caption{Simulation executions, cases S01--S07 from top to bottom. Task names and instructions appear above each strip; frames progress from left to right.}
\label{fig:simulation}
\end{figure}

\begin{figure}[H]
\centering
\includegraphics[width=\linewidth]{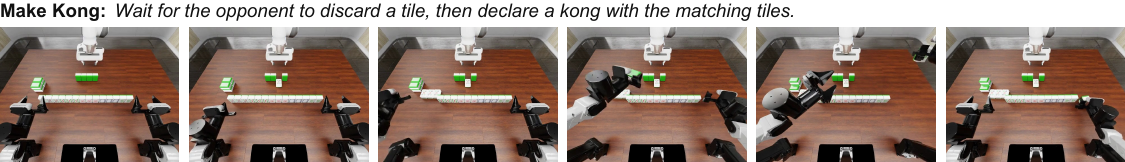}
\par\vspace{6pt}
\includegraphics[width=\linewidth]{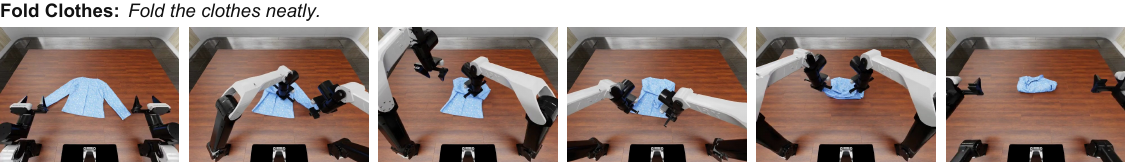}
\par\vspace{6pt}
\includegraphics[width=\linewidth]{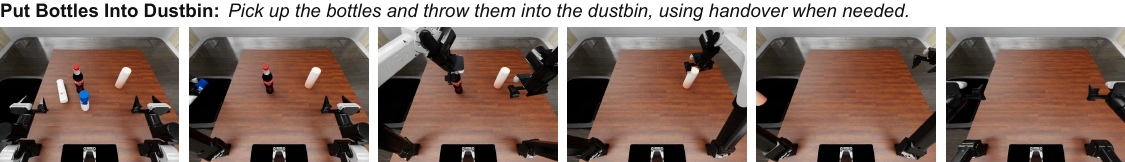}
\par\vspace{6pt}
\includegraphics[width=\linewidth]{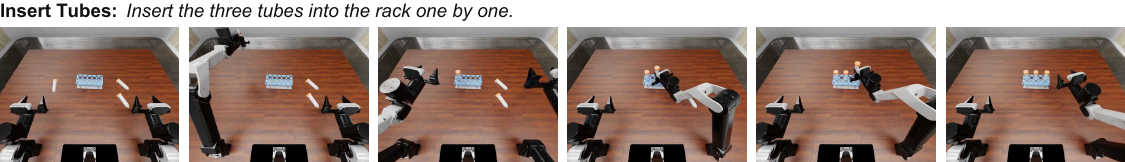}
\par\vspace{6pt}
\includegraphics[width=\linewidth]{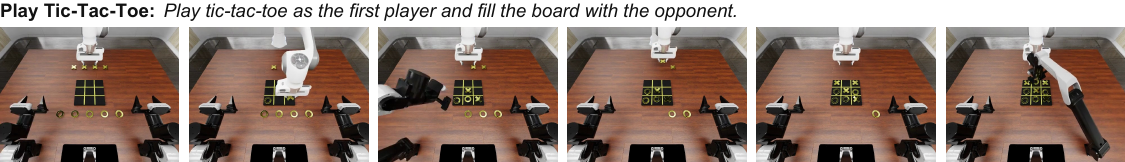}
\par\vspace{6pt}
\includegraphics[width=\linewidth]{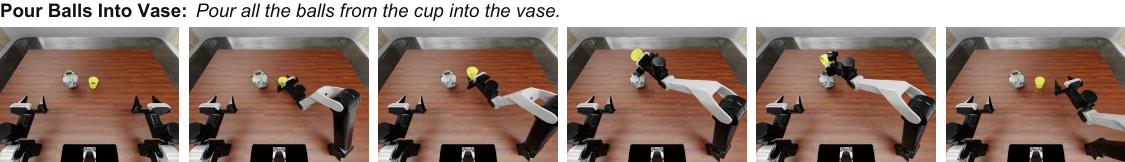}
\par\vspace{6pt}
\includegraphics[width=\linewidth]{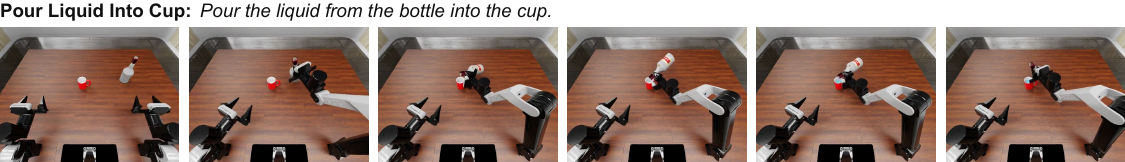}
\caption{Simulation executions, cases S08--S14 from top to bottom. Task names and instructions appear above each strip; frames progress from left to right.}
\label{fig:simulation_s08_s14}
\end{figure}

\begin{figure}[H]
\centering
\includegraphics[width=\linewidth]{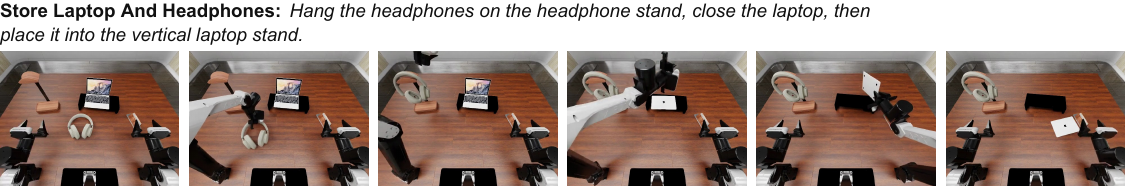}
\par\vspace{6pt}
\includegraphics[width=\linewidth]{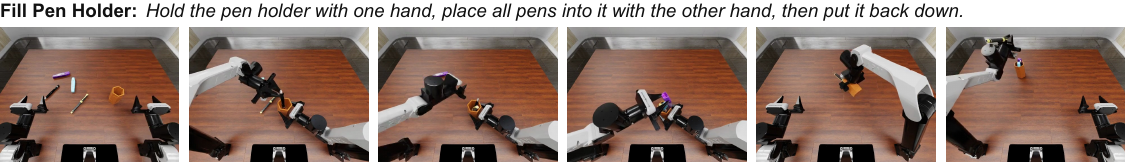}
\par\vspace{6pt}
\includegraphics[width=\linewidth]{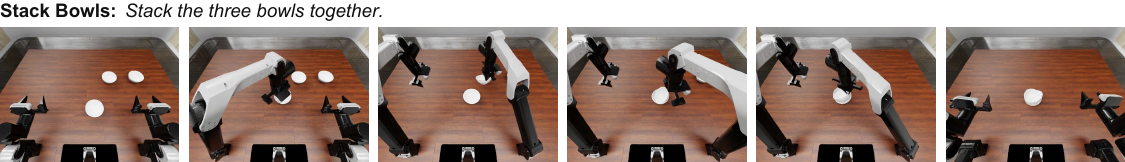}
\par\vspace{6pt}
\includegraphics[width=\linewidth]{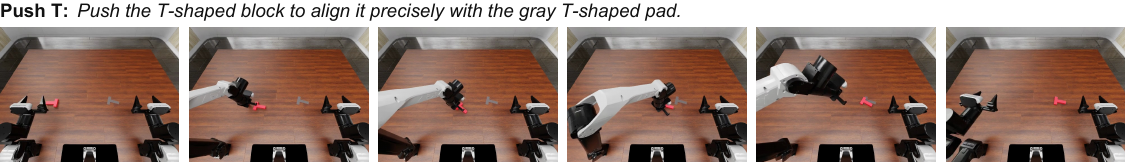}
\par\vspace{6pt}
\includegraphics[width=\linewidth]{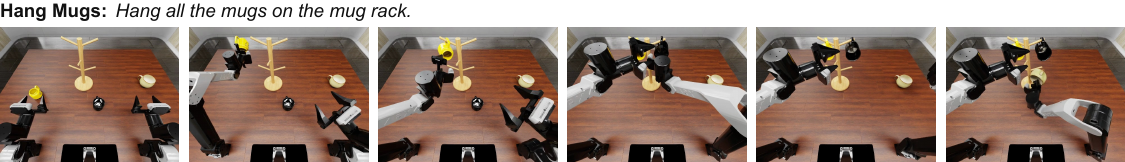}
\par\vspace{6pt}
\includegraphics[width=\linewidth]{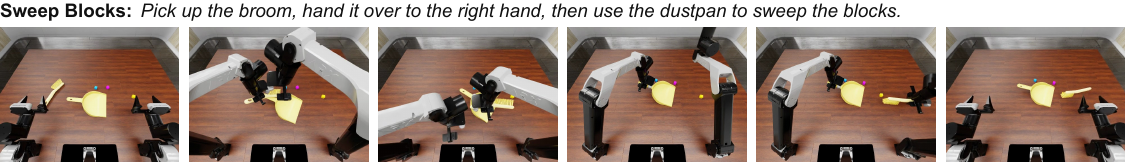}
\par\vspace{6pt}
\includegraphics[width=\linewidth]{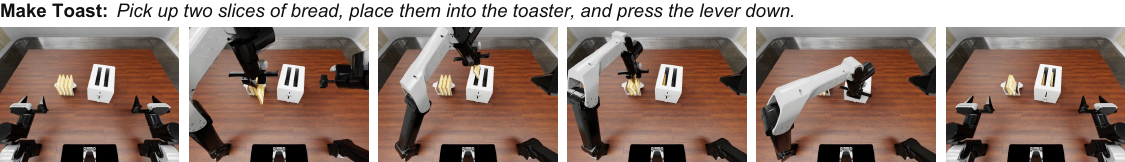}
\caption{Simulation executions, cases S15--S21 from top to bottom. Task names and instructions appear above each strip; frames progress from left to right.}
\label{fig:simulation_s15_s21}
\end{figure}

\begin{figure}[H]
\centering
\includegraphics[width=\linewidth]{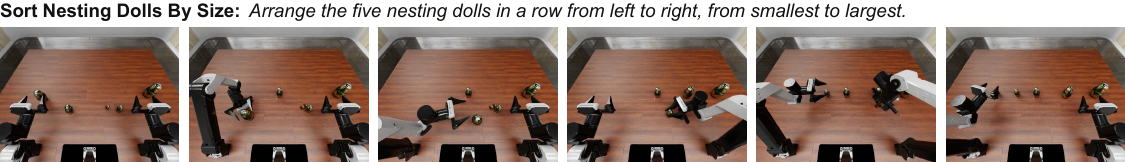}
\par\vspace{6pt}
\includegraphics[width=\linewidth]{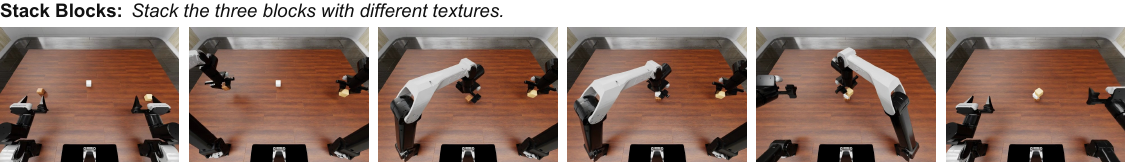}
\par\vspace{6pt}
\includegraphics[width=\linewidth]{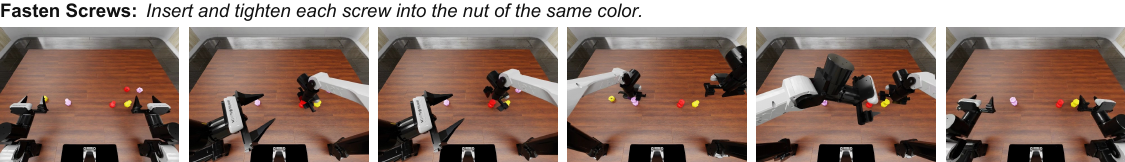}
\par\vspace{6pt}
\includegraphics[width=\linewidth]{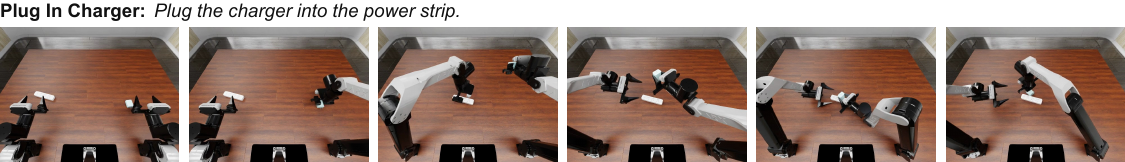}
\par\vspace{6pt}
\includegraphics[width=\linewidth]{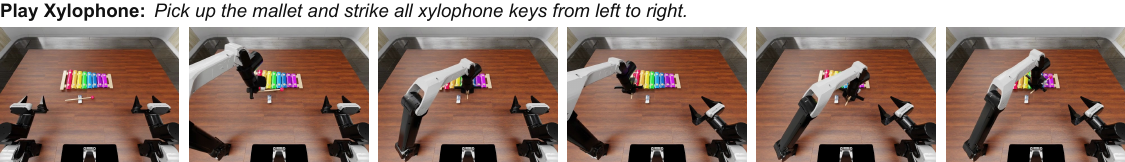}
\par\vspace{6pt}
\includegraphics[width=\linewidth]{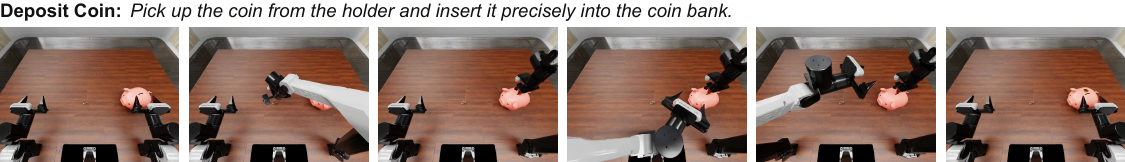}
\par\vspace{6pt}
\includegraphics[width=\linewidth]{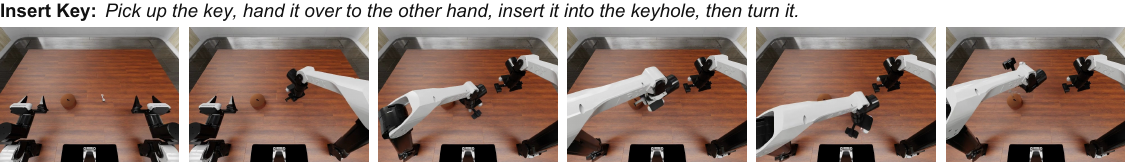}
\caption{Simulation executions, cases S22--S28 from top to bottom. Task names and instructions appear above each strip; frames progress from left to right.}
\label{fig:simulation_s22_s28}
\end{figure}

\begin{figure}[H]
\centering
\includegraphics[width=\linewidth]{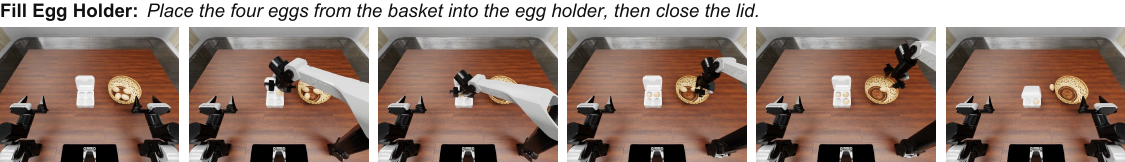}
\par\vspace{6pt}
\includegraphics[width=\linewidth]{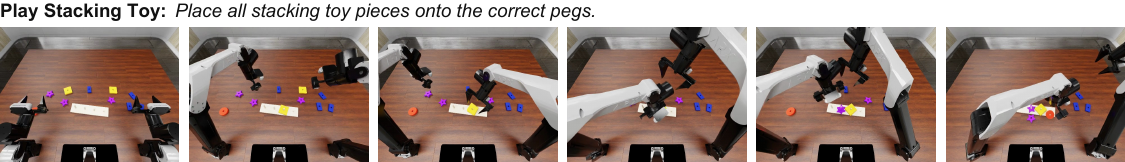}
\par\vspace{6pt}
\includegraphics[width=\linewidth]{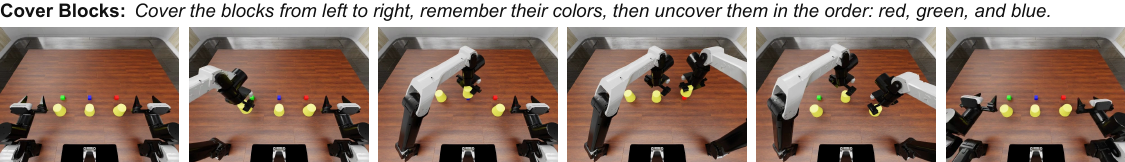}
\par\vspace{6pt}
\includegraphics[width=\linewidth]{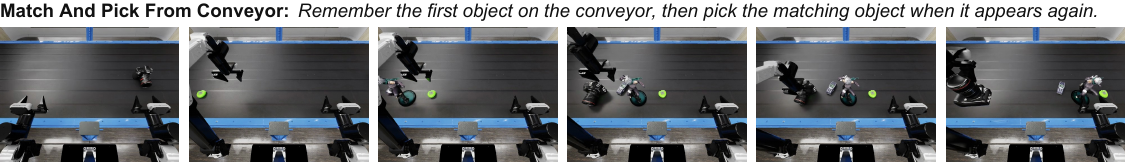}
\par\vspace{6pt}
\includegraphics[width=\linewidth]{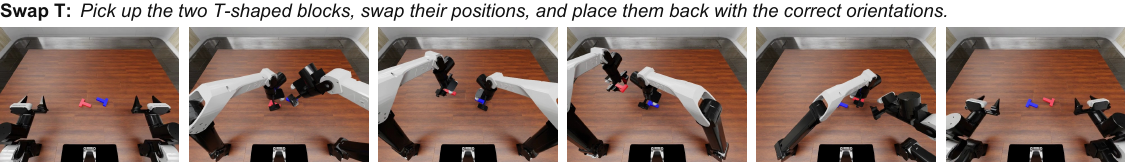}
\par\vspace{6pt}
\includegraphics[width=\linewidth]{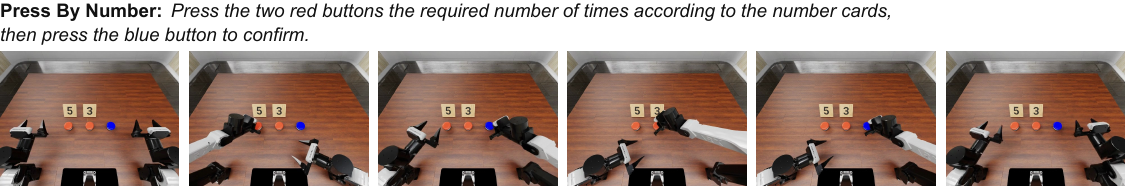}
\par\vspace{6pt}
\includegraphics[width=\linewidth]{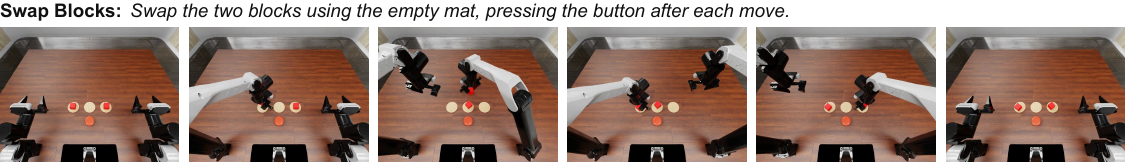}
\caption{Simulation executions, cases S29--S35 from top to bottom. Task names and instructions appear above each strip; frames progress from left to right.}
\label{fig:simulation_s29_s35}
\end{figure}

\begin{figure}[H]
\centering
\includegraphics[width=\linewidth]{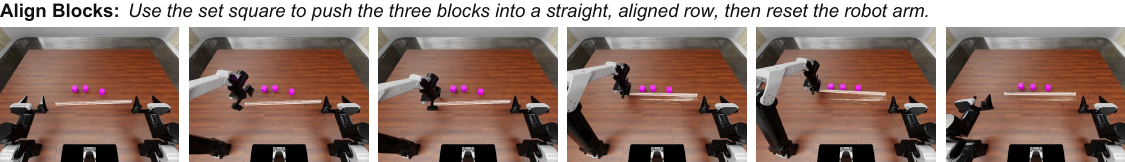}
\par\vspace{6pt}
\includegraphics[width=\linewidth]{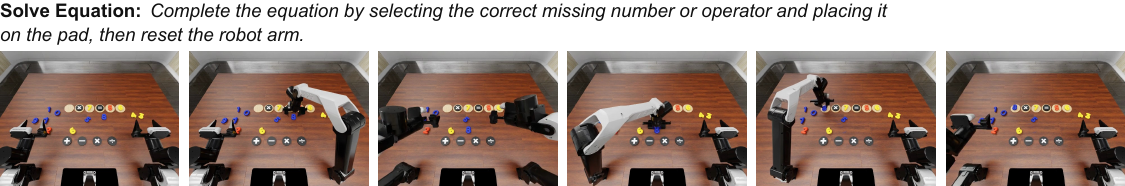}
\par\vspace{6pt}
\includegraphics[width=\linewidth]{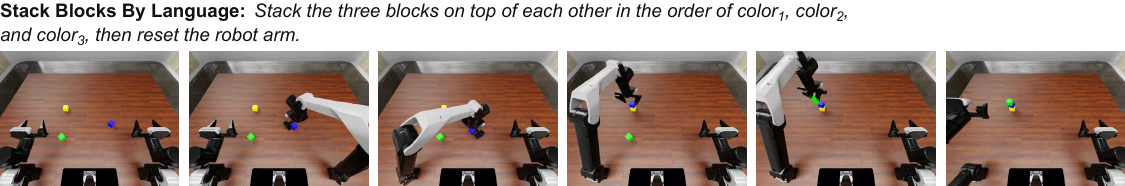}
\par\vspace{6pt}
\includegraphics[width=\linewidth]{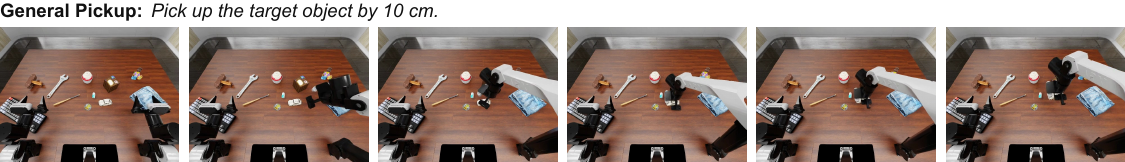}
\par\vspace{6pt}
\includegraphics[width=\linewidth]{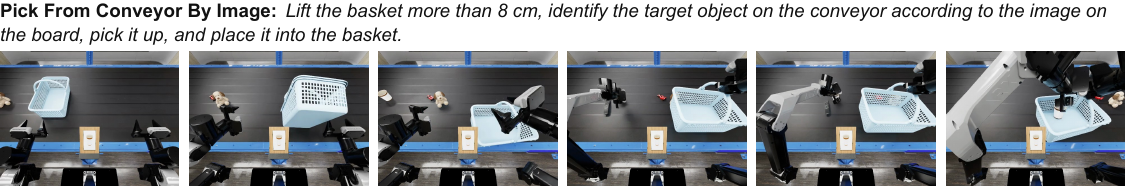}
\par\vspace{6pt}
\includegraphics[width=\linewidth]{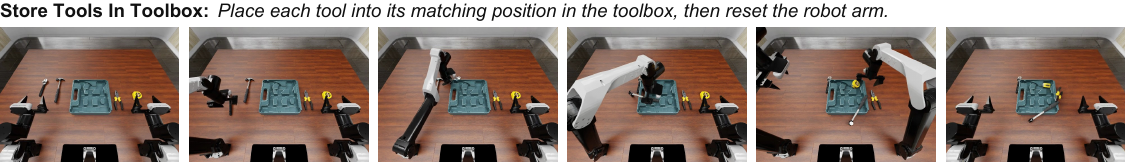}
\par\vspace{6pt}
\includegraphics[width=\linewidth]{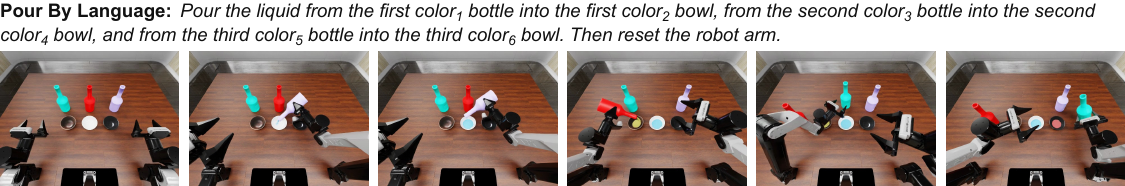}
\caption{Simulation executions, cases S36--S42 from top to bottom. Task names and instructions appear above each strip; frames progress from left to right.}
\label{fig:simulation_s36_s42}
\end{figure}

\Cref{fig:simulation,fig:simulation_s08_s14,fig:simulation_s15_s21,fig:simulation_s22_s28,fig:simulation_s29_s35,fig:simulation_s36_s42} present execution sequences for all 42 RoboDojo simulation tasks. Each strip retains the task name and instruction and shows six frames in temporal order from left to right. The gallery follows the S01--S42 case order.

\clearpage

\clearpage

\clearpage

\clearpage

\clearpage

\clearpage
\begingroup\raggedright
\section{Task-Agent Input, Output, and Tool Formats}
\label{app:taskinterface}
The following blocks describe the task-agent interface in a joint-control configuration. Angle brackets and ellipses denote runtime data rather than literal argument values.

\subsection{Input Context}
\label{app:inputformat}
The system message supplies the control contract and selected skills. Each skill is Markdown text with \texttt{name} and \texttt{description} fields. A request also includes the preceding completed tool cycle, current public context, and RGB observations. Key context fields are:
\begingroup\fontsize{10}{12.5}\selectfont
\begin{verbatim}
{
  "instruction": "<current task instruction>",
  "step": <native-step index>,
  "state": {
    "left_arm_joint_state": [<six measured joint angles>],
    "right_arm_joint_state": [<six measured joint angles>],
    "left_ee_pose": [<x>, <y>, <z>, <qw>, <qx>, <qy>, <qz>],
    "right_ee_pose": [<x>, <y>, <z>, <qw>, <qx>, <qy>, <qz>],
    "left_ee_joint_state": [<gripper command echo>],
    "right_ee_joint_state": [<gripper command echo>],
    "left_gripper_measured_joints": [<measured finger positions>],
    "right_gripper_measured_joints": [<measured finger positions>]
  },
  "memory": "<same-episode facts and unresolved questions>",
  "previous_result": <preceding tool feedback or null>,
  "budget": {
    "remaining_native_steps": <integer>,
    "model_calls_remaining": <integer>, ...
  }, ...
}
\end{verbatim}
\endgroup
Named RGB inputs include \texttt{cam\_head}, \texttt{cam\_left\_wrist}, and \texttt{cam\_right\_wrist}, together with selected previous-segment or requested historical views. Other context fields provide home poses, joint semantics, and camera geometry. Joint angles are in radians; positions are in metres.

\paragraph{Task and progress fields.}
\texttt{instruction} is the natural-language goal for the current task. \texttt{step} is the current native environment-step index: one motion call may advance several native steps, whereas a geometry or review call need not advance it. \texttt{state} groups the robot's reported measurements at that step.

\paragraph{Arm-state fields.}
\texttt{left\_arm\_joint\_state} and \texttt{right\_arm\_joint\_state} contain six measured joint angles for the corresponding arm. \texttt{left\_ee\_pose} and \texttt{right\_ee\_pose} give the reported end-effector position and orientation. In each seven-element pose, the first three entries are $x,y,z$ in metres; the final four are the quaternion components in scalar-first order $q_w,q_x,q_y,q_z$.

\Needspace{6\baselineskip}
\paragraph{Gripper-state fields.}
\texttt{left\_ee\_joint\_state} and \texttt{right\_ee\_joint\_state} echo gripper commands. \texttt{left\_gripper\_measured\_joints} and \texttt{right\_gripper\_measured\_joints} instead report measured finger positions. The distinction lets the agent compare what was commanded with what the gripper physically reached; neither a closing command nor a small opening by itself proves that an object is held.

\Needspace{8\baselineskip}
\paragraph{Memory, feedback, and budgets.}
\texttt{memory} carries the task agent's explicit same-episode notes, including established facts and unresolved questions. It is not a persistent skill document. \texttt{previous\_result} carries the preceding tool's feedback, or \texttt{null} when no preceding result is available. \texttt{budget.remaining\_native\_steps} is the remaining environment-step allowance, while \texttt{budget.model\_calls\_remaining} is the remaining model-call allowance. These are separate resources because a tool call can execute zero, one, or several native steps.

\Needspace{6\baselineskip}
\paragraph{Skills and image names.}
A skill's \texttt{name} identifies the document and its \texttt{description} summarizes its purpose or applicability; its Markdown body supplies procedural guidance. The named \texttt{cam\_head}, \texttt{cam\_left\_wrist}, and \texttt{cam\_right\_wrist} inputs identify the source camera for RGB observations. RGB observations are supplied as named image inputs alongside the JSON context.

\Needspace{29\baselineskip}
\subsection{Model Output and Function Calls}
\label{app:outputformat}
The model returns exactly one enabled function call. Every tool requires \texttt{assessment}, \texttt{plan}, \texttt{phase}, and updated episode \texttt{memory}, followed by tool-specific arguments. Below, \texttt{arguments} is displayed as a decoded object; the function-call transport serializes it as a JSON string.
\begingroup\fontsize{10}{12.5}\selectfont
\begin{verbatim}
{
  "type": "function_call", "call_id": "<call identifier>",
  "name": "control_arms",
  "arguments": {
    "assessment": "<observed evidence and uncertainty>",
    "plan": "<next operation>",
    "phase": "fine",
    "memory": "<updated same-episode memory>",
    "left": null,
    "right": {
      "control": "joint",
      "joints": [<q1>, <q2>, <q3>, <q4>, <q5>, <q6>],
      "gripper": null
    },
    "steps": <integer>, "settle": <integer>
  }
}
\end{verbatim}
\endgroup

\Needspace{5\baselineskip}
\paragraph{Call-envelope fields.}
\texttt{type: function\_call} marks an executable tool request. \texttt{call\_id} identifies that request and must match the identifier in its return message. \texttt{name} selects an enabled tool, here \texttt{control\_arms}; \texttt{arguments} contains the fields validated against that tool's schema.

\Needspace{5\baselineskip}
\paragraph{Common decision fields.}
\texttt{assessment} summarizes the currently observed evidence and remaining uncertainty. \texttt{plan} states the next intended operation. \texttt{phase} selects the fine-motion or transit regime and its permitted segment length. The output \texttt{memory} is the updated note carried into the next decision.

\Needspace{7\baselineskip}
\paragraph{Arm-selection fields.}
\texttt{left} and \texttt{right} select commands for the corresponding arms. A non-null arm object uses \texttt{control: joint} to select joint control, \texttt{joints} for its target vector, and \texttt{gripper} for its opening command. An arm or an individual target field set to \texttt{null} preserves that command. \texttt{steps} requests the native motion-segment length, while \texttt{settle} requests additional hold steps for settling. Units and allowed ranges are detailed below.

\Needspace{24\baselineskip}
\subsection{Tool Arguments and Execution Feedback}
\label{app:toolformat}
\Cref{tab:tasktools} lists tool-specific arguments in addition to the four common fields above. Only enabled tools can be invoked; task-agent calls do not provide shell access or hidden simulator-state queries.

\begin{table}[H]
\centering
\caption{Task-agent tool arguments. Braced entries denote object fields, and square brackets denote lists.}
\label{tab:tasktools}
\begingroup\small
\setlength{\tabcolsep}{4pt}\renewcommand{\arraystretch}{1.14}
\begin{tabularx}{\linewidth}{@{}P{0.34\linewidth}L@{}}
\toprule
Tool & Tool-specific arguments and semantics \\
\midrule
\texttt{control\_arms} & \texttt{left}, \texttt{right}, \texttt{steps}, \texttt{settle}. Each non-null arm has \texttt{control}, \texttt{joints}, and \texttt{gripper} fields, as shown above. \\
\texttt{review\_frames} & \texttt{frames: [\{step, camera\}]}. Retrieves 1--6 same-episode RGB frames and their public poses without motion. \\
\texttt{project\_pixels} & \texttt{points: [\{pixel: [u,v], z\}]}. Projects 1--32 head-view pixels onto model-specified height planes without motion. \\
\texttt{triangulate\_point} & \texttt{step}, \texttt{feature}, \texttt{views: [\{camera, pixel\}]}. Matches one physical feature in 2--3 distinct current views; returns geometry estimates and diagnostics without motion. \\
{\fontsize{10}{12.5}\selectfont\texttt{triangulate\_temporal\_point}} & \texttt{step}, \texttt{camera}, \texttt{feature}, \texttt{stationary\_target}, \texttt{camera\_hand\_empty}, \texttt{stationarity\_evidence}, and \texttt{samples: [\{step, pixel\}]}. Uses 2--3 reviewed views from one wrist camera. Both Boolean fields must be \texttt{true}; sampling steps increase, end at the current step, and span at most 24 native steps. \\
\texttt{verify\_completion} & No additional arguments. Advances six hold steps and checks native termination. \\
\texttt{stop\_episode} & \texttt{reason}. Stops the controller when blocked. \\
\bottomrule
\end{tabularx}
\endgroup
\end{table}

\Needspace{8\baselineskip}
\paragraph{Observation and geometry fields.}
In \texttt{review\_frames}, \texttt{frames} is a list of retrieval requests: \texttt{step} selects a same-episode observation by its native-step index and \texttt{camera} selects its view. In geometric queries, \texttt{pixel: [u,v]} denotes horizontal and vertical image coordinates. The \texttt{z} in \texttt{project\_pixels} specifies a model-chosen height plane in metres; it is a hypothesis, not a supplied object depth. In \texttt{triangulate\_point}, \texttt{step} identifies the current observation, \texttt{feature} describes the same physical feature to match, and \texttt{views} pairs each camera with its pixel coordinates.

\Needspace{8\baselineskip}
\paragraph{Temporal geometry and stopping fields.}
For \texttt{triangulate\_temporal\_point}, \texttt{camera} selects one wrist camera, \texttt{feature} identifies the target feature, and \texttt{samples} pairs observation steps with its pixel locations. \texttt{stationary\_target} and \texttt{camera\_hand\_empty} state the required stationary-target and empty-camera-hand conditions; \texttt{stationarity\_evidence} explains their observational basis. These flags must be supported by the observations. The \texttt{reason} argument of \texttt{stop\_episode} explains why execution should stop.

\Needspace{7\baselineskip}
\paragraph{Control parameters.}
Each non-null \texttt{joints} field contains six absolute targets in radians, not joint increments. \texttt{gripper} is a continuous opening command in $[0,1]$. A null arm, joint vector, or gripper field holds its corresponding command. \texttt{phase} is \texttt{fine} or \texttt{transit}: motion segments use 1--6 or 1--20 native steps, respectively, with \texttt{settle} in $\{0,1,2\}$. Execution checks can stop a segment before the requested number of steps.

\Needspace{23\baselineskip}
\paragraph{Tool-result format.}
The result is linked to the model call through the same \texttt{call\_id}. For a motion tool, feedback includes segment status, the executed step count, and tracking measurements. The following abbreviated block displays the serialized \texttt{output} payload as a decoded object:
\begingroup\fontsize{10}{12.5}\selectfont
\begin{verbatim}
{
  "type": "function_call_output",
  "call_id": "<same call identifier>",
  "output": {
    "feedback": {
      "status": "<segment status>",
      "reason": "<status detail>",
      "executed_steps": <integer>,
      "tracking": [<per-arm tracking measurements>], ...
    },
    "step": <current native-step index>,
    "ended": <Boolean>
  }
}
\end{verbatim}
\endgroup
The next request supplies updated context and images together with this feedback. Observation and geometry tools instead return their requested evidence or diagnostic information without executing native actions.

\Needspace{5\baselineskip}
\paragraph{Return-envelope fields.}
\texttt{type: function\_call\_output} identifies a tool return. Its \texttt{call\_id} links it to the preceding request, and \texttt{output} contains the tool's returned payload. As with \texttt{arguments}, the block displays the decoded content rather than the transport's JSON string.

\Needspace{7\baselineskip}
\paragraph{Execution-feedback fields.}
\texttt{feedback.status} describes the motion segment's execution status, such as completion, interruption, or rejection. \texttt{feedback.reason} provides an associated explanation and may be \texttt{null} when no detail applies. \texttt{feedback.executed\_steps} reports how many native steps actually ran, which can differ from the requested motion and settling steps. \texttt{feedback.tracking} contains per-arm measurements comparing commanded and observed execution.

\Needspace{5\baselineskip}
\paragraph{Updated step and termination.}
The return's \texttt{step} is the updated native-step index after the tool call. \texttt{ended} indicates whether the episode has terminated, while task success is determined separately by the environment's native evaluator. Updated state, images, and feedback allow the next call to respond to what actually happened.

\par
\endgroup

\end{document}